\pdfoutput=1

\documentclass[11pt]{article}

\usepackage[]{acl}

\usepackage{times}
\usepackage{latexsym}
\usepackage{hyperref}
\usepackage{url}
\usepackage{multirow}
\usepackage{booktabs}
\usepackage{amssymb}
\usepackage{titlesec}
\usepackage{booktabs}
\usepackage{enumerate}
\usepackage[font=small,labelfont=bf]{caption}
\usepackage{graphicx}
\usepackage{listings}
\usepackage{pythonhighlight}
\usepackage{longtable}
\usepackage{inconsolata}
\usepackage{xspace}
\usepackage{hyperref}
\usepackage{wrapfig,lipsum}
\newcommand{\ours}{NLEP\xspace}

\usepackage[T1]{fontenc}

\usepackage[utf8]{inputenc}

\usepackage{microtype}

\usepackage{inconsolata}

\title{Natural Language Embedded Programs for\\ Hybrid Language Symbolic Reasoning}

\author{Tianhua Zhang$^{\dagger*}$, Jiaxin Ge$^{\diamondsuit*}$, Hongyin Luo$^\ddagger$\thanks{$\;\;$Equal contribution. Correspondence to Hongyin Luo at \url{hyluo@mit.edu}. Work done during Jiaxin Ge visiting MIT. Source code is available at \url{https://github.com/luohongyin/LangCode}.}$\;\;$, Yung-Sung Chuang$^\ddagger$, Mingye Gao$^\ddagger$,\\ \bf Yuan Gong$^\ddagger$, Xixin Wu$^\dagger$, Yoon Kim$^\ddagger$, Helen Meng$^\dagger$, James Glass$^\ddagger$ \\
$^\dagger$The Chinese University of Hong Kong, $^\ddagger$Massachusetts Institute of Technology, $^\diamondsuit$Peking University\\
\texttt{tzhang@se.cuhk.edu.hk, aomaru@stu.pku.edu.cn, hyluo@mit.edu}\\
}

\begin{document}
\maketitle
\begin{abstract}
How can we perform  computations over natural language representations to solve tasks that require symbolic and numeric reasoning? We propose \emph{natural language embedded programs} (\ours) as a unifying framework for  addressing math/symbolic reasoning, natural language understanding, and instruction following tasks. Our approach prompts a  language model to generate full Python programs that define functions over data structures which contain natural language representations of structured knowledge. A Python interpreter then executes the generated code and prints the output. Despite using a task-general prompt, we find that this approach can improve upon strong baselines across a range of different tasks including math and symbolic reasoning, text classification, question answering, and instruction following. We found that the generated programs are interpretable since they outline the exact reasoning process followed by the program interpreter.
\end{abstract}

\section{Introduction}
\begin{figure*}[h]
\centering
\includegraphics[width=.8\textwidth]{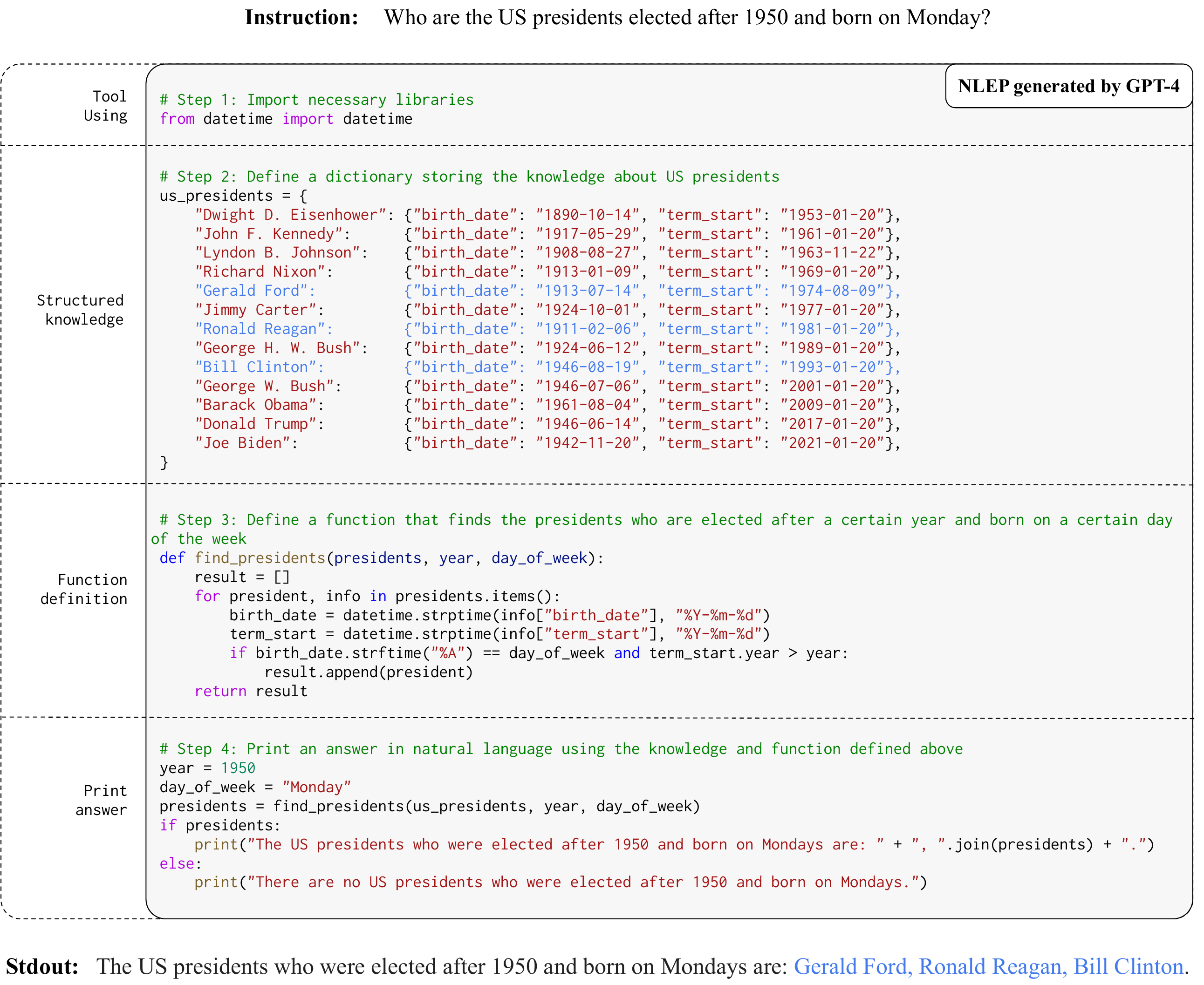}
\caption{A generated NLEP correctly answers the given question while ChatGPT-4 obtains an incorrect  answer (\href{https://chat.openai.com/share/524cb7bb-b6f4-4a70-9ba9-4f7b17f23c3e}{link}). This NLEP uses the date-weekday conversion tool in the \texttt{datetime} package, constructs structured knowledge about US presidents, implements a selection function, and outputs natural language responses depending on the function output. A more detailed comparison between NLEP and ChatGPT-4 code interpreter is shown in Figure \ref{fig:figure1}.
}
\label{fig:prompt}
\end{figure*}

Solving complex language tasks often requires performing  computations over natural language representations. For language-based reasoning, chain-of-thought  prompting~\citep[CoT;][]{wei2022chain} has emerged as a promising approach for surfacing the symbolic  reasoning capabilities of large language models (LLMs). However, certain types of computations (e.g., arithmetic) are unnatural to perform in pure language space, and hence present difficulties for LLMs. General-purpose programming languages, on the other hand, provide convenient abstractions as well as predefined libraries and functions for natively implementing many types of symbolic computations, and there has been much recent work on interleaving program calls  within CoT-style reasoning to extend the capabilities of LLMs. While promising, existing methods are generally limited to narrow types of tasks such as math and symbolic reasoning \citep{chen2022program,cai2023large,gao2023pal}, simple API calling \citep{schick2023toolformer,paranjape2023ART,liang2023taskmatrix}, and database accessing \citep{cheng2022binding}. These works moreover rely on task-specific prompts which are hard to generalize across datasets.

This work describes a task-general approach for combining the language-based reasoning capabilities of LLMs with symbolic computations enabled by the use of programs. Specifically, we prompt LLMs to generate \emph{natural language embedded programs} (NLEPs), which are fully executable Python programs containing appropriate package importing, structured natural language representations of knowledge, function definitions for problem solving, and response printing. The generated NLEP is then executed using a Python interpreter that captures the standard output of the program as the response. An example of an NLEP generated by GPT-4 is shown in Figure \ref{fig:prompt}.

NLEPs use code as a scaffold to reason over  natural language representations of data and knowledge. This makes our approach different from ToolFormer \citep{schick2023toolformer} and language model as tool maker \citep[LATM;][]{cai2023large}, which instead use language as the scaffold and interleave API calls within natural language sentences during LLM generation. Compared to program-of-thought \citep[PoT;][]{chen2022program} and program aided language models \citep[PAL;][]{gao2023pal}, which mainly focus on math and symbolic problems, NLEPs handles natural language, math, and symbolic tasks with significantly better generalization ability based on task-general prompts. Compared to existing neuro-symbolic augmentations approaches needing new logical language \citep{li2023scallop} and predefined solvers \citep{pan2023logic}, NLEP uses a general programming interface thus can handle a significantly wider range of tasks with more flexible programming elements including packages, databases, and APIs.

Experiments across math and symbolic reasoning, question answering and instruction following, and text classification tasks demonstrate that (1) NLEPs conducts accurate reasoning on both structured and unstructured tasks and inputs; (2) NLEP's step-by-step, meta prompting strategy can significantly improve the prompt efficiency across different tasks. As a result, we conclude that programming language prompting with NLEP is more capable and generalizable than existing natural language and neuro-symbolic prompting strategies.
 

\vspace{-2mm}
\section{Approach: NLEP Prompting}
\vspace{-2mm}
\label{section: code-of-thought-prompting}


An NLEP is a program containing both programming code and natural language. NLEPs use natural language in several different ways. First, it uses natural language comments to guide step-by-step program generation. Second, language is used to represent structured knowledge through Python's native data structures (e.g., dictionaries and lists). Finally, an NLEP uses language to print fluent responses to the user input by constructing a standard output string containing references to program variables. 

The hybrid language-symbolic design of \ours enables generalized problem solving for natural language, math, symbolic reasoning, and API calling tasks, which have traditionally been  tackled by separate mechanisms.  This approach combines the benefits of language-based reasoning with program synthesis: comments and knowledge in natural language improve program generation, while the structured/symbolic reasoning powered by program interpreters provides more accurate computations than would have been obtained via direct decoding from LLMs.

An example of an NLEP for answering a question is shown in Figure \ref{fig:figure1}. In the generated program, each section is preceded by comments in natural language, and the defined counting function uses knowledge stored in a key-value dictionary (which itself is generated from GPT-4's internal knowledge) to find the correct answer. Finally, the answer is printed through a natural language response. In this example, we generated 5 independent NLEPs and found that they achieve 100\% accuracy, compared to 60\% for ChatGPT-4 and 40\% GPT-4 API.

\textbf{NLEP structure.} More generally, each NLEP contains four sections: (1) importing necessary libraries, (2) defining variables containing structured knowledge, (3) implementing problem-solving functions, and (4) printing the response in natural language. Instead of providing direct solutions for each task, we guide the model to arrive at a solution following this four-step process. The structured and modularized programming design disentangles the knowledge extraction and reasoning steps, in contrast to PoT \citep{chen2022program}, where different segments may intertwine. 
As show in the example in Figure \ref{fig:prompt}, an NLEP answers the question by constructing a structured knowledge dictionary containing the birthday and start date of the US presidents. To recognize the weekdays, the program utilizes pre-defined functions in the \texttt{datetime} package. The selected answers are stored in a \texttt{list} and then embedded into an output template. The \ours also handles the situation when no answer is found. The correct answer is then printed by the \ours.

\textbf{Task-general demonstration prompts.}
As is standard in chain-of-thought prompting \citep{nye2021work,wei2022chain}, our approach  uses demonstration prompts for NLEP generation. However, unlike previous approaches our demonstrations are not task-specific. The unified NLEP structure motivates models to solve different problems in the same disentangled four steps, bypassing the necessity for task-specific, in-domain examples to explicitly teach the models for a particular task. For example, for all classification tasks  we consider we use the \emph{same} demonstration prompt (derived from SST2). Similarly, we use mostly the same prompt for our math and symbolic reasoning tasks. This task-general prompt is similar in spirit to {zero-shot chain-of-thought} prompting \citep{kojima2023large} which adds a task-agnostic prompt (``\texttt{Let's think step-by-step}'') to elicit the reasoning capabilities of LLMs in a task-agnostic way. The  prompts used for the various tasks are given in Table \ref{table:datasets}, and the exact prompts are given in Appendix \ref{appendix-1-prompt}. In summary, we use 4 different demonstration prompts across 16 tasks, each of which works well within a task category. Thus, while the proposed method is not fully task-agnostic in the strictest sense of the term, it  is still more flexible than previous approaches that combine program synthesis with chain-of-thought prompting \citep{chen2022program,gao2023pal}, which use examples from the dataset to craft prompts.

\begin{figure*}[t]
\centering
\includegraphics[width=.9\textwidth]{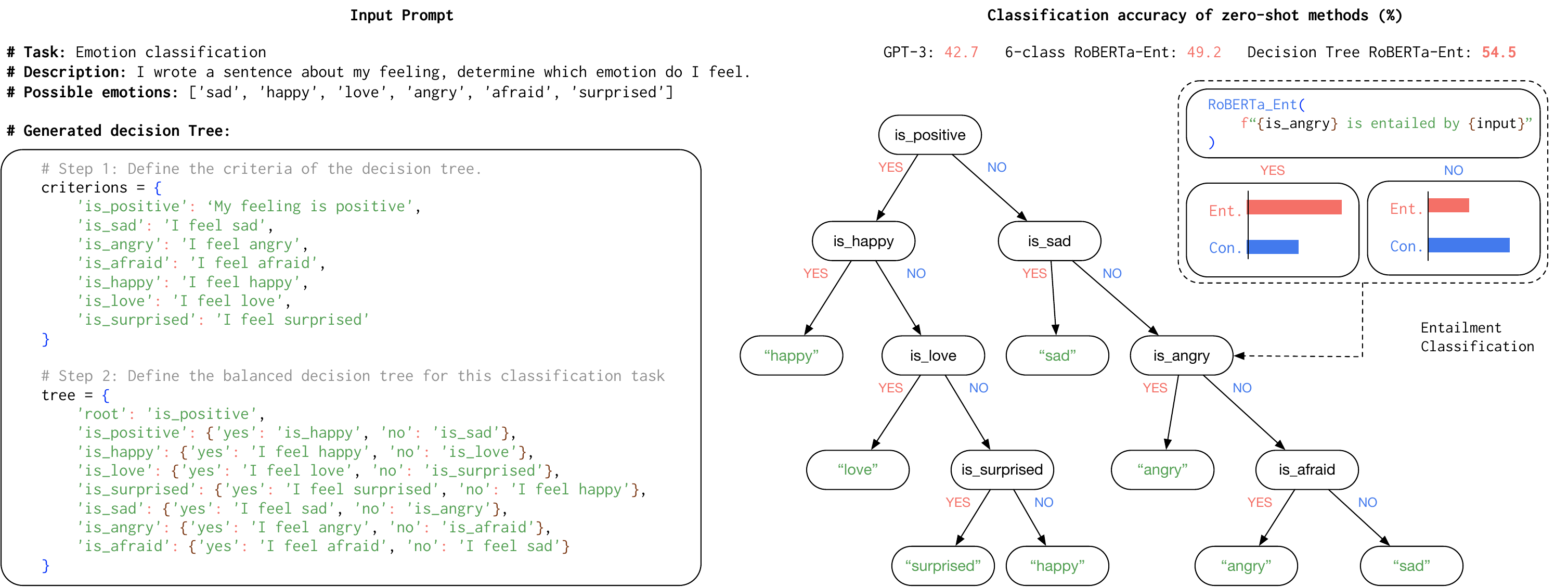}
\caption{A decision tree structure generated within an \ours for emotion classification based on task description using an example program for SST2 as the prompt. The branching of each node is decided by a RoBERTa \citep{liu2019roberta} text entailment model. This language-based decision tree generated by an NLEP outperforms GPT-3 and entailment-based multi-class prediction \citep{ge2023entailment} without needing any task-specific examples (i.e., exemplars specific to the emotion classification dataset).
}
\label{fig:tree}
\end{figure*}

\textbf{Programmatic reasoning for natural language understanding.} 
Prior works on combining program synthesis with LLM-based reasoning have generally focused on  math and symbolic reasoning tasks \citep{chen2022program,gao2023pal}, and it has not been clear how such methods could be extended to address natural language understanding (NLU) tasks. We show that NLEPs can be straightforwardly extended to text classification tasks.

For  question answering, we apply \ours prompting and the target output is constructed by the generated programs. Classification tasks, on the other hand, are handled by a different type of NLEP consisting of a decision tree since standard program-based reasoning may not seamlessly translate to text-based classification and yield optimal performance. Each node of the decision tree is annotated by a simple natural language sentence, and the \texttt{Yes}/\texttt{No} decisions at each node are handled in a zero-shot way by an entailment classifier,   which has in general been shown to be an effective approach to zero-shot text classification \citep{obamuyide-vlachos-2018-zero,condoravdi-etal-2003-entailment,ge2023entailment}. Concretely, given the tree we compute the entailment score between the input and the language description associated with each node and traverse the decision tree until a leaf node is reached. We emphasize that the topology of the tree and the textual description of each node is generated by the prompted LLM only \textit{once} per benchmark. The decision maker per question is indeed the smaller entailment models for tree traversal. Consequently, \ours leverages LLMs to efficiently construct decision trees, assisting smaller models in making informed decisions and augmenting overall performance. The demonstration prompt for classification tasks is given by a manually constructed example from SST2 \citep{wang2018glue} for different benchmarks. We find that this prompt can generate NLEPs containing sensible decision trees for various classification tasks without requiring task-specific examples. An example of the generated program and the corresponding decision tree is shown in Figure \ref{fig:tree}.

\vspace{-2mm}
\section{Experiments}
\vspace{-2mm}
We evaluate  natural language embedded programs (NLEPs)  on 16 tasks across three broad task categories. The tasks and corresponding prompts are summarized  in Table \ref{table:datasets}.
 
\textbf{Math and symbolic reasoning} tasks include Tracking Shuffled Objects, Dyck Language, Word Sorting and Chinese Remainder Theorem from BigBench \citep{srivastava2023imitation}, Scheduling Meeting task from \citet{cai2023large}, GSM-Hard\footnote{We opted for GSM-Hard to mitigate potential data contamination from GSM8K. The numbers in GSM-Hard are large, less likely to have been encountered by LLMs during training, and hence augmenting the task complexity. However, since GSM-Hard is automatically constructed, 
there are cases where the answers are not reasonable or the questions appear peculiar. We detail the analysis in Appendix \ref{appendix-3-additional-symbolic}.} benchmark of math word problems from \citet{gao2023pal}, and Game of 24 \citep{yao2023tree}. 
We use two examples for all tasks except for Game of 24, for which we applied a word sorting example to elicit stronger game-playing reasoning ability. The exact \ours prompts we used are given in Appendix \ref{appendix:p4t26} and \ref{appendix:p4t3}.

\textbf{Question answering and instruction following} tasks include the StrategyQA \citep{geva2021did}, TruthfulQA \citep{lin2022truthfulqa}, and VicunaQA \citep{vicuna2023} benchmarks. StrategyQA requires models to answer multi-hop questions with ``\texttt{Yes}'' or ``\texttt{No}''. TruthfulQA and VicunaQA contain questions and instructions requiring free-form responses. VicunaQA also allows us to test how NLEPs perform in the popular instruction-following setting. The evaluation metrics on question answering focus on accuracy, relevance, and factuality of the generated answers. The prompts in Appendix \ref{appendix:p4t26} are used for StrategyQA. For TruthfulQA and VicunaQA, we added an example with a longer response to encourage more detailed response generation.

\textbf{Text classification} tasks includes tasks that require understanding of both natural language inputs and labels. We evaluate \ours on movie-review classification \citep[SST2;][]{socher-etal-2013-recursive}, linguistic acceptance \citep[COLA;][]{warstadt-etal-2019-neural}, emotion classification \citep{saravia-etal-2018-carer}, amazon review \citep{ni2019justifying}, hate speech detection \citep{de-gibert-etal-2018-hate}, and stereotypes recognition \citep{sap2019social}. We use the prompts in Appendix \ref{appendix:p4t26} for model-free classification. For decision tree generation, the prompts in Appendix \ref{appendix:p4t45} are applied.

\begin{table}[]
\centering
\scalebox{0.7}{
\begin{tabular}{llc}
\toprule
Domain                                                                                                             & Task                       & Prompt    \\ \midrule
\multicolumn{1}{c}{\multirow{6}{*}{\begin{tabular}[c]{@{}l@{}}Math and Symbolic\\Reasoning\end{tabular}}}                                                           & Tracking Shuffled Objects (7) & \ref{appendix:p4t26}     \\
\multicolumn{1}{c}{}                                                                                              & Dyck Language                      & \ref{appendix:p4t26} \\
\multicolumn{1}{c}{}                                                                                              & Word Sorting                       & \ref{appendix:p4t26} \\
\multicolumn{1}{c}{}                                                                                              & Chinese Remainder Theorem          & \ref{appendix:p4t26}     \\
\multicolumn{1}{c}{}                                                                                              & Scheduling Meeting                 & \ref{appendix:p4t26} \\
\multicolumn{1}{c}{}                                                                                              & GSM-Hard   & \ref{appendix:p4t26}     \\ 
\multicolumn{1}{c}{} & Game of 24  & \ref{appendix:p4t3} 
\\

\midrule
\multicolumn{1}{l}{\multirow{3}{*}{Question Answering}}                                                                            & StrategyQA                    & \ref{appendix:p4t26}                   \\
& TruthfulQA  & \ref{appendix:p4t7f5}  \\
& VicunaQA  & \ref{appendix:p4t7f5}  \\
\midrule

\multicolumn{1}{l}{\multirow{6}{*}{\begin{tabular}[c]{@{}l@{}}Text Classification \\ \end{tabular}}} &    SST2        &  \ref{appendix:p4t45}             \\
\multicolumn{1}{l}{}                                                                                              &            Cola               &          \ref{appendix:p4t45}                    \\
\multicolumn{1}{l}{}                                                                                              &       Emotion-Classification                      &        \ref{appendix:p4t45}                        \\
\multicolumn{1}{l}{}                                                                                              &        Amazon Review            &            \ref{appendix:p4t45}                 \\ 
\multicolumn{1}{l}{}                                                                                              &       Hate-Speech                         &        \ref{appendix:p4t45}                        \\
\multicolumn{1}{l}{}                                                                                              &       Social Bias Frame                  &        \ref{appendix:p4t45}                        \\
\bottomrule
\end{tabular}}
\caption{Summary descriptions of the various tasks considered in this work.}
\label{table:datasets}
\vspace{-0.2cm}
\end{table}

\subsection{Math and Symbolic Reasoning}


We compare  \ours prompting with chain-of-thought \citep[CoT;][]{wei2022chain}, program-of-thought  \citep[PoT;][]{chen2022program}, and LLMs as tool makers \citep[LATM;][]{cai2023large}.
We also compare against tree-of-thought  \citep[ToT;][]{yao2023tree} on the Game of 24 benchmark, where ToT  outperforms CoT by a significant margin  (but requires many more calls to the LLM). 
We evaluate CoT and PoT with both task-general and task-specific demonstrations. Since LATM needs in-domain input-output pairs to create tools, we only report the results with task-specific LATM. 

\textbf{Task-general prompting.} For task-general prompts we use two examples as the in-context demonstration for the math and symbolic reasoning benchmarks (see Table~\ref{table:datasets} and Appendix \ref{appendix-1-prompt}).
For CoT, we present two examples with intermediate reasoning represented in natural language rather than as programs.
Our task-general PoT implementation takes the math and symbolic reasoning lines similar as \cite{chen2022program} and \cite{gao2023pal}, but without the step-by-step programming scheme in \ours as an ablation. 

\textbf{Task-specific prompting baselines.} We report the task-specific prompting performance as an ``upper bound'' for each task. For CoT, we use the same prompting settings (from 3 to 8-shot) adopted in previous studies \citep{cobbe2021training,cai2023large,fu2023chainofthought}.
For PoT, we use the same in-context examples as in the task-specific CoT examples, but provide intermediate reasoning steps in Python code. On the GSM-Hard benchmark, we adopt the demonstrations (9-shot) for GSM8K used in \cite{chen2022program}. For the Chinese Remainder Theorem and Scheduling Meeting benchmarks, we construct the in-context examples with the first three successful instances of task-general PoT.  
For LATM, we evaluate its performance on Tracking Shuffled Objects (7) using the provided tool 
and cite the results for other tasks from \cite{cai2023large}. Details are shown in Appendix \ref{appendix-2-fewshot-symbolic}.

Program synthesis approaches (PoT and \ours) may sometimes generate non-executable programs if lack task-specific programming demonstration. For both approaches, we select certain benchmarks to resample up to three additional programs if the returned program failed at execution.
Since this condition is triggered only if program execution fails, there is no label leakage. We discuss this further in Section \ref{sec-discussion} and provide results details in Appendix \ref{appendix-3-additional-symbolic}.
\begin{table*}[h]
\centering
\renewcommand{\arraystretch}{1.2}
\scalebox{0.7}{
\begin{tabular}{l|cccccc|ccccc}
\toprule
{\color[HTML]{000000} }                                          & \multicolumn{6}{c|}{{\color[HTML]{000000} \textbf{GPT-4}}}                                                                                                                                                                                                                                                      & \multicolumn{5}{c}{{\color[HTML]{000000} \textbf{GPT-3.5-Turbo}}}                                                                                                                                                                                                       \\ \cline{2-12} 
{\color[HTML]{000000} }                                          & \multicolumn{3}{c|}{{\color[HTML]{656565} \textbf{(a) Task-Specific}}}                                                                         & \multicolumn{3}{c|}{{\color[HTML]{000000} \textbf{(b) Task-General}}}                                                                                                 & \multicolumn{2}{c|}{{\color[HTML]{656565} \textbf{(c) Task-Specific}}}                               & \multicolumn{3}{c}{{\color[HTML]{000000} \textbf{(d) Task-General}}}                                                                                                    \\
\multirow{-3}{*}{{\color[HTML]{000000} \textbf{Tasks / Method}}} & {\color[HTML]{656565} \textbf{CoT}}   & {\color[HTML]{656565} \textbf{PoT}}   & \multicolumn{1}{c|}{{\color[HTML]{656565} \textbf{LATM}}}  & {\color[HTML]{000000} \textbf{CoT}} & {\color[HTML]{000000} \textbf{PoT}} & {\color[HTML]{000000} \textbf{\begin{tabular}[c]{@{}c@{}}\ours 
\end{tabular}}} & {\color[HTML]{656565} \textbf{CoT}} & \multicolumn{1}{c|}{{\color[HTML]{656565} \textbf{PoT}}}   & {\color[HTML]{000000} \textbf{CoT}} & {\color[HTML]{000000} \textbf{PoT}}   & {\color[HTML]{000000} \textbf{\begin{tabular}[c]{@{}c@{}}\ours 
\end{tabular}}} \\ \midrule
{\color[HTML]{000000} Tracking Shuffled Objects}             & {\color[HTML]{656565} \textbf{100.0}} & {\color[HTML]{656565} \textbf{100.0}} & \multicolumn{1}{c|}{{\color[HTML]{656565} \textbf{100.0}}} & {\color[HTML]{000000} 81.2}         & {\color[HTML]{000000} 98.4}         & {\color[HTML]{000000} \textbf{100.0}}                                                  & {\color[HTML]{656565} 68.0}         & \multicolumn{1}{c|}{{\color[HTML]{656565} 6.8}}            & {\color[HTML]{000000} 51.2}         & {\color[HTML]{000000} 88.4}           & {\color[HTML]{000000} 74.4}                                                            \\
{\color[HTML]{000000} Dyck Language}                             & {\color[HTML]{656565} 63.6$^\dagger$}           & {\color[HTML]{656565} 60.8}           & \multicolumn{1}{c|}{{\color[HTML]{656565} 87.5$^\dagger$}}           & {\color[HTML]{000000} 39.6}         & {\color[HTML]{000000} 66.4}         & {\color[HTML]{000000} \textbf{91.6}}                                                   & {\color[HTML]{656565} 20.4$^\dagger$}         & \multicolumn{1}{c|}{{\color[HTML]{656565} 28.4}}           & {\color[HTML]{000000} 38.0}         & {\color[HTML]{000000} 4.0}            & {\color[HTML]{000000} 7.2}                                                             \\
{\color[HTML]{000000} Word Sorting}                              & {\color[HTML]{656565} 90.9$^\dagger$}           & {\color[HTML]{656565} \textbf{100.0}} & \multicolumn{1}{c|}{{\color[HTML]{656565} 99.1$^\dagger$}}           & {\color[HTML]{000000} 84.4}         & {\color[HTML]{000000} 99.6}         & {\color[HTML]{000000} 99.6}                                                            & {\color[HTML]{656565} 59.2$^\dagger$}         & \multicolumn{1}{c|}{{\color[HTML]{656565} \textbf{100.0}}} & {\color[HTML]{000000} 75.2}         & {\color[HTML]{000000} \textbf{100.0}} & {\color[HTML]{000000} 99.6}                                                            \\
{\color[HTML]{000000} Chinese Remainder Theorem}                 & {\color[HTML]{656565} 0.0$^\dagger$}            & {\color[HTML]{656565} \textbf{100.0}} & \multicolumn{1}{c|}{{\color[HTML]{656565} \textbf{100.0$^\dagger$}}} & {\color[HTML]{000000} 0.0}          & {\color[HTML]{000000} 84.4}         & {\color[HTML]{000000} 97.2}                                                            & {\color[HTML]{656565} 0.0$^\dagger$}          & \multicolumn{1}{c|}{{\color[HTML]{656565} \textbf{100.0}}} & {\color[HTML]{000000} 0.0}          & {\color[HTML]{000000} 72.4}           & {\color[HTML]{000000} 96.4}                                                            \\
{\color[HTML]{000000} Scheduling Meeting}                        & {\color[HTML]{656565} 55.6$^\dagger$}           & {\color[HTML]{656565} 75.2}           & \multicolumn{1}{c|}{{\color[HTML]{656565} \textbf{100.0$^\dagger$}}} & {\color[HTML]{000000} 82.8}         & {\color[HTML]{000000} 85.2}         & {\color[HTML]{000000} 93.2}                                                            & {\color[HTML]{656565} 18.9$^\dagger$}         & \multicolumn{1}{c|}{{\color[HTML]{656565} 33.6}}           & {\color[HTML]{000000} 39.6}         & {\color[HTML]{000000} 49.2}           & {\color[HTML]{000000} 85.6}                                                            \\
{\color[HTML]{000000} GSM-Hard}                                  & {\color[HTML]{656565} 57.4}           & {\color[HTML]{656565} \textbf{74.1}}  & \multicolumn{1}{c|}{{\color[HTML]{656565} --}}              & {\color[HTML]{000000} 54.9}         & {\color[HTML]{000000} 69.3}         & {\color[HTML]{000000} 67.7}                                                            & {\color[HTML]{656565} 45.0}         & \multicolumn{1}{c|}{{\color[HTML]{656565} 63.4}}           & {\color[HTML]{000000} 42.8}         & {\color[HTML]{000000} 52.2}           & {\color[HTML]{000000} 54.1}                                                            \\ \midrule
{\color[HTML]{000000} \textbf{Average}}  & {\color[HTML]{656565} 61.3} & {\color[HTML]{656565} 85.0} &  \multicolumn{1}{c|}{{\color[HTML]{656565} \textbf{97.3}}} & 57.2 & 83.9 & \multicolumn{1}{c|}{91.6} & {\color[HTML]{656565} 35.3} & \multicolumn{1}{c|}{\color[HTML]{656565} {55.4}} & 41.1 & 61.0 & 69.6\\
\bottomrule
\end{tabular}}
\caption{
Performance on math and symbolic reasoning tasks with both task-specific and task-general demonstration prompts. $^\dagger$ stands for results from \cite{cai2023large}. LATM results are not available for GSM-Hard benchmark as it is hard to derive a generally applicable tool function for all test cases.
}
\label{table:symbolic-reasoning}
\vspace{-0.2cm}
\end{table*}

\subsubsection{Results}

\begin{figure*}[t]
\centering
\includegraphics[width=.9\textwidth]{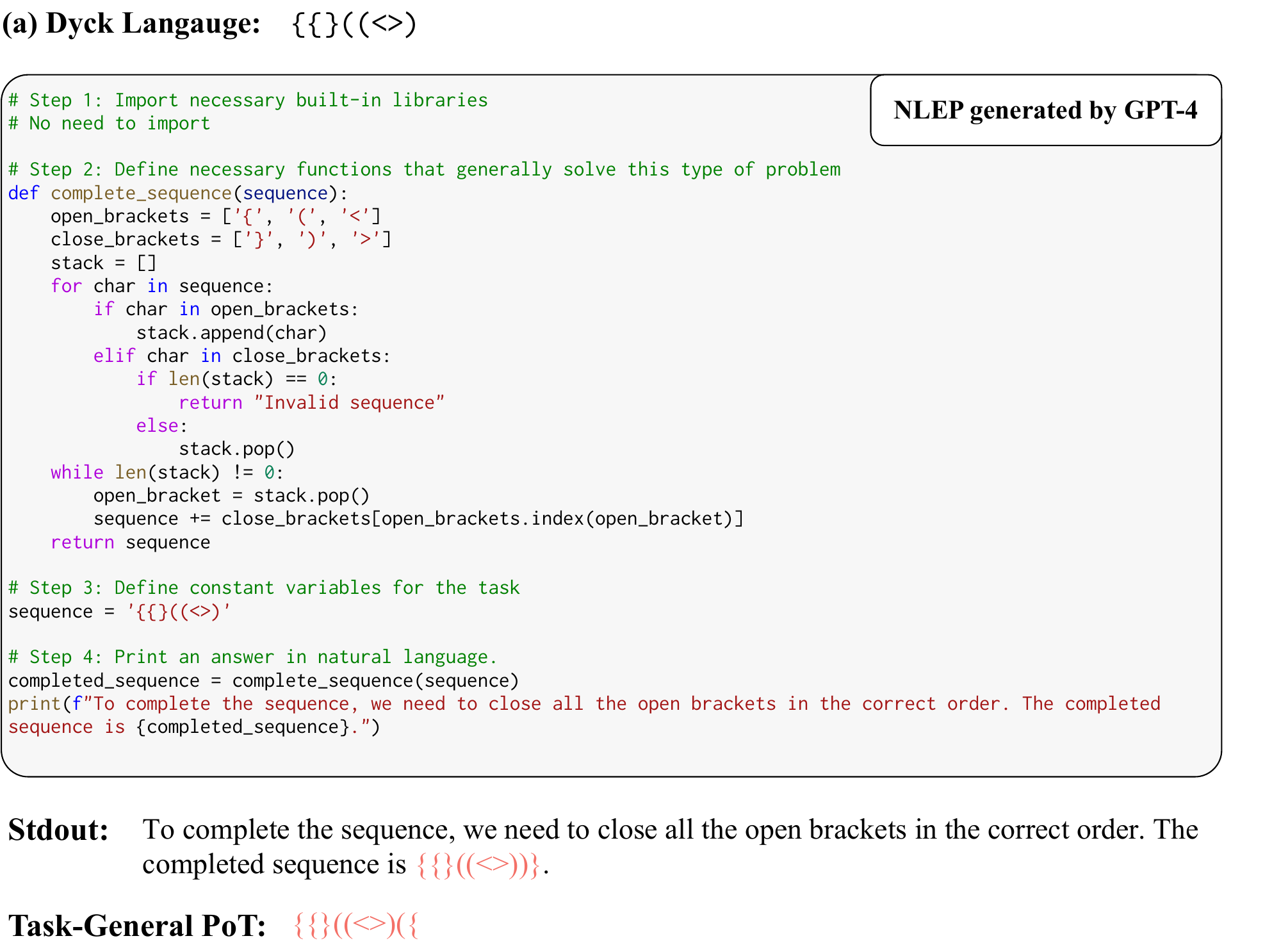}
\caption{\ours generated for solving Dyck language problem. The instruction is \textit{``Complete the rest of the sequence, making sure that the parentheses are closed properly.''} An example for StrategyQA is outlined in Figure \ref{fig:sqa-example}.
}
\label{fig:sqa-dyck}
\end{figure*}

We show the main results of NLEP prompting on six math and symbolic reasoning tasks in Table \ref{table:symbolic-reasoning}. An example of \ours generated for solving a Dyck language problem is shown in Figure \ref{fig:sqa-dyck}(a).

\textbf{GPT-4 Results.}
Among the three approaches employing task-general prompts, \ours demonstrates superior performance over CoT across 5 of 6 tasks and outperforms PoT on 4 of 6 tasks. Additionally, \ours achieves equivalent performance to PoT on Word Sorting benchmark.
The large performance gap between \ours and CoT suggests that programmatic reasoning can enable more accurate answers.
Compared to PoT, \ours achieves significantly higher average accuracy, especially on the Dyck Language (66.4\%$\rightarrow$91.6\%)
and the complex Chinese Remainder Theorem (84.4\%$\rightarrow$97.2\%) tasks. On GSM-Hard, we confirm the same phenomenon discovered by  \cite{gao2023pal} where language does not further benefit the calculation accuracy with GPT-4. 

\ours also achieves comparable performance to task-specific, few-shot prompting methods.
Notably, our method achieves the best performance on Tracking Shuffled Objects (7) and Dyck Language, and outperforms task-specific CoT on many benchmarks.
On the Word Sorting benchmark, \ours only fails on one instance where the input word sequence contains ``steelmake" and GPT-4 automatically  corrected it to ``steelmaker". 
We find that the high scores of task-specific PoT on Word Sorting and Chinese Remainder Theorem come from the generally applicable programming code from the in-context demonstrations.


\textbf{GPT-3.5 Results.} 
We observe a significant decline in performance with GPT-3.5 for all reasoning approaches across nearly all benchmarks.
However \ours still achieves the best average performance,
exhibiting significant improvement on 5 of 6 tasks over both task-specific and task-general CoT baselines. \ours exceeds task-general PoT notably across 4 tasks and demonstrates comparable performance on the Word Sorting benchmark.
On the Dyck Language benchmark, program-based strategies (PoT and \ours with task-general prompts) failed to accomplish the problem without task-specific examples, highlighting the need for strong backbone LLMs.

\begin{table}[h]
\small
\centering
\begin{tabular}{@{}llc@{}}
\toprule
\textbf{Prompt}                & \multicolumn{1}{c}{\textbf{Method}} & \textbf{Accuracy (\%)} \\ \midrule
\multirow{3}{*}{Task-specific} & CoT                                 & 4                      \\
                               & ToT (b = 1)                         & 45                     \\
                               & ToT (b = 5)                         & 74                     \\ \midrule
\multirow{2}{*}{Task-general}  & PoT                                 & 52                     \\
                               & NLEP                                & 66                     \\ \bottomrule
\end{tabular}
\caption{Performance on the Game of 24 benchmark. CoT and ToT stand for chain-of-thought \citep{wei2022chain} and tree-of-thought \citep{yao2023tree} prompting respectively. $^\dagger$ shows the results from \cite{yao2023tree}.} 
\label{table:game24}
\vspace{-0.2cm}
\end{table}

\textbf{Game of 24 results.} Table \ref{table:game24} shows the results  on the challenging Game of 24 task from \cite{yao2023tree}.
Our approach also surpasses the oracle setup of IO/CoT, which calculates the success rate of IO/CoT by considering the best of 100 samples for each instance. However, unlike ToT which requires in-context demonstrations for each decomposed sub-task,
\ours prompting achieves a significant performance gain over ToT (b=1) without requiring a computationally expensive multi-chain inference procedure.

\subsection{QA and Instruction Following}

\begin{table*}[h]
\renewcommand{\arraystretch}{1.2}
\centering
\scalebox{0.8}{
\begin{tabular}{c|cccc|cccc}
\hline
\multirow{3}{*}{setting}        & \multicolumn{4}{c|}{GPT-4}                                              & \multicolumn{4}{c}{GPT-3.5-Turbo}                                      \\ \cline{2-9} 
                                & \multicolumn{1}{c|}{Task-specific} & \multicolumn{3}{c|}{Task-general} & \multicolumn{1}{c|}{Task-specific} & \multicolumn{3}{c}{Task-general} \\
                                & \multicolumn{1}{c|}{CoT}           & CoT        & PoT       & \ours (ours)      & \multicolumn{1}{c|}{CoT}           & CoT       & PoT       & \ours (ours)      \\ \hline
\multicolumn{1}{l|}{StrategyQA} & \multicolumn{1}{c|}{\textbf{81.7}} & 78.6       & 68.6      & 81.2      & \multicolumn{1}{c|}{71.6} & 68.1      & 50.2      & 61.1      \\ \hline
\end{tabular}}
\caption{Performance on the StrategyQA benchmark. The experimental setup is the same as in Table \ref{table:symbolic-reasoning}. 
Note that LLMs do not always generate ``\texttt{Yes}'' or ``\texttt{No}''. and we only predict the ``\texttt{Yes}'' label if the ``\texttt{Yes}'' string is generated explicitly. 
See Appendices \ref{appendix:p4t26} and \ref{appendix-2-fewshot-symbolic} for implementation details.}
\label{table:strategyqa}
\end{table*}

\textbf{StrategyQA.} Experiment results are presented in Table \ref{table:strategyqa}. With GPT-4, \ours achieves the best performance under the task-general prompt setting and is competitive with the task-specific CoT. 
With GPT-3.5, although the scores of code-based strategies decrease more than CoT (PoT: 18.4\%, NLEP: 20.1\%, task-general CoT: 10.5\%, task-specific CoT: 10.1\%), \ours still exceeds PoT by a significant margin. An example of output is shown in \ref{fig:sqa-dyck}(b).

\begin{table}
\centering
\scalebox{0.83}{
\begin{tabular}{@{}lcccc@{}}
\toprule
\textbf{Foundation Model}      & \textbf{Mode} & \textbf{True} & \textbf{Info} & \textbf{True * Info} \\ \midrule
\multirow{2}{*}{GPT-4}         & Text          & \textbf{76.01}         & 97.55         & 73.56                \\
                               & \ours         & 75.76         & \textbf{99.63}         & \textbf{75.40}                \\ \midrule
\multirow{2}{*}{GPT-3.5-Turbo} & Text          & 68.91         & 98.90         & 67.93                \\
                               & \ours         & 61.69         & 97.18         & 59.00                \\ \bottomrule
\end{tabular}
}
\caption{Performance of GPT-4 and GPT-3.5-Turbo on the TruthfulQA benchmark.}
\label{table:tqa}
\vspace{-0.2cm}
\end{table}

\textbf{TruthfulQA.} We also evaluate how \ours prompting influences the factuality of question answering with the TruthfulQA benchmark \citep{lin2022truthfulqa}. A fine-tuned GPT-3 model is applied for automatic scoring. In this experiment, we compare the vanilla auto-regressive text generation method against \ours. Traditionally, such question answering tasks have been solved only with black-box language model without explicit symbolic computations due to the complexity of  test questions.

The results are shown in Table \ref{table:tqa}. With  GPT-4, the truth score of \ours prompting strategy is close to standard LLM-based generation, while the informativeness score is higher. However, performance degrades significant with  GPT-3.5-Turbo, indicating a strong dependence on the programming ability of the underlying language model.

\begin{figure}[h]
  \centering
  \begin{minipage}[c]{0.48\textwidth}
    \centering
    \scalebox{0.83}{
    \begin{tabular}{@{}lcccc@{}}
    \toprule
    \small
    \textbf{Model}      & \textbf{\small \# \ours\textgreater Text} & \textbf{\small Detail} & \textbf{\small \% Score} & \parbox{15mm}{\centering \textbf{\small \% Length \\ Bias}} \\ \midrule
    \multirow{2}{*}{GPT-4}         & \multirow{2}{*}{23.75}                   & yes             & 93.08                   & 72.72                   \\
                                   &                                          & no              & \textbf{105.06}         & 26.67                   \\ \midrule
    GPT-3.5 & \multirow{2}{*}{38.75}                   & yes             & 101.22                  & \textbf{3.13}           \\
                           -Turbo        &                                          & no              & 102.50                  & 10.34                   \\ \bottomrule
    \end{tabular}}
  \end{minipage}
  \hfill
  \begin{minipage}[c]{0.48\textwidth}
    \includegraphics[width=\textwidth]{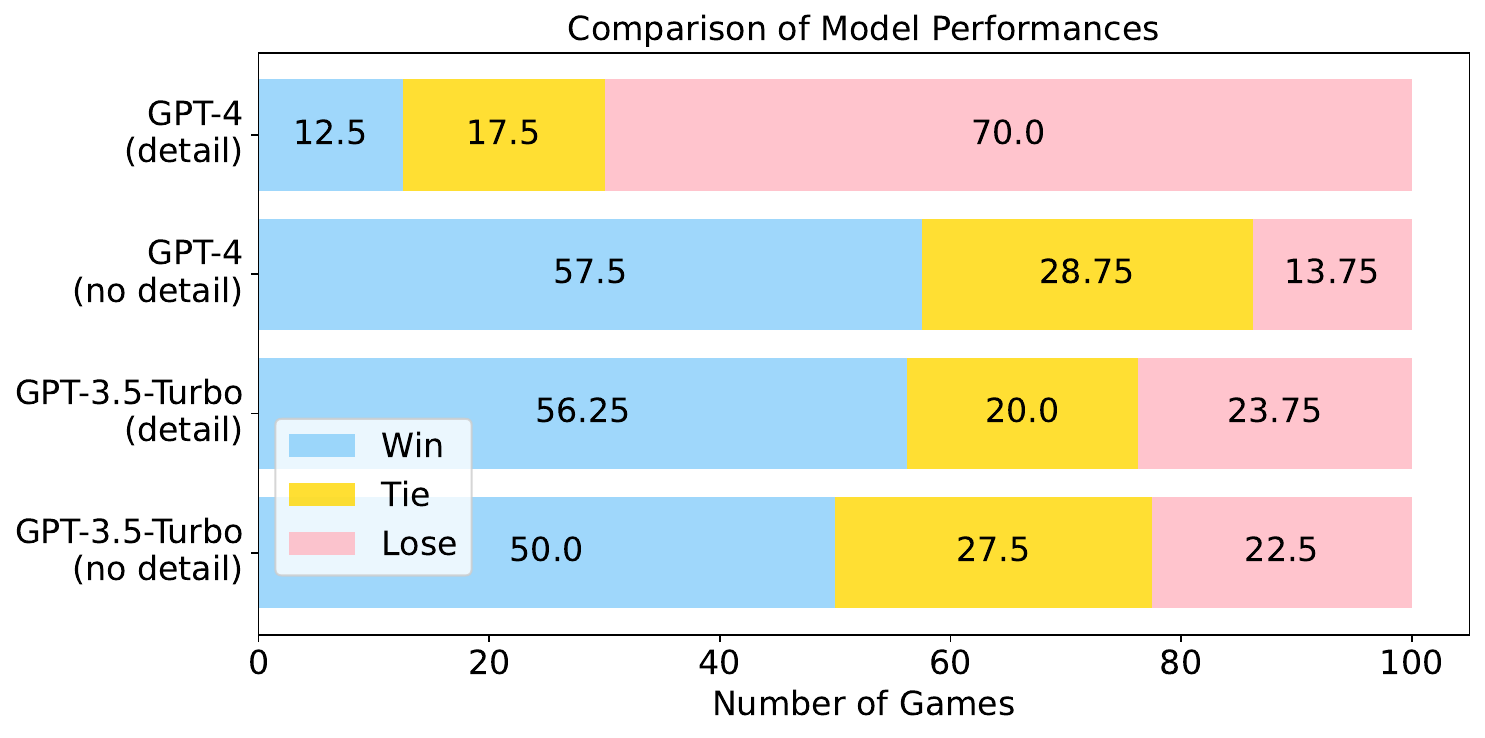}
  \end{minipage}
  \caption{Automatic evaluations of \ours against standard LLM-based generation with different  models. \textbf{\# \ours \textgreater Text} means that the \% of \ours responses containing more tokens than the baseline. \textbf{Detail} means if the evaluation metric considers details and response lengths. \textbf{Score} stands for the scores received by \ours divided by the baseline scores (\textgreater 100 means \ours is better). \textbf{Win}, \textbf{tie}, and \textbf{lose} stand for the \% of evaluation cases resulting in each category. \textbf{Length Bias} shows how much the evaluation pipeline prefers longer or shorter answers (lower means fairer, introduced in Appendix \ref{appendix:lb}).}
  \label{table:vicunaqa}
\end{figure}

\textbf{VicunaQA.} Results on the VicunaQA benchmark are shown in Figure \ref{table:vicunaqa}, where we follow the standard approach and evaluate the answers using GPT-4. We find that GPT-4 prefers its own generations, which are generally more detailed than GPT-3.5-Turbo and \ours responses. To control for the bias due to response lengths, we also assess all responses without the requirement about details using another evaluation prompt. The evaluation prompts with and without the requirement on details is shown in Appendix \ref{appendix:prompt-detail} and \ref{appendix:prompt-no-detail}.

As we demonstrate in Figure \ref{table:vicunaqa}, this assessment leads to different results on GPT-4. After removing the detail requirement in the automatic scoring pipeline,  \ours  achieves better performance. This suggests that \ours can help GPT-4 generate accurate, factual, and relevant responses. However, human-generated programs for pretraining the GPT-4 models usually do not embed long pieces of natural language. As a result, the responses generated by \ours have a limited level of detail. 

\subsection{Text Classification}
Finally, we evaluate whether NLEPs can be applied to solve text classification tasks that have traditionally been difficult for pure program synthesis-based approaches. As discussed in section~\ref{section: code-of-thought-prompting}, we manually construct a decision tree \ours for SST2 and use it as a prompt to guide GPT models to generate decision trees for other tasks only with task and label descriptions. We highlight that the decision tree is generated \textit{once} per benchmark. Subsequently, all decision-making for each specific question is carried out with RoBERTa/DeBERTa to perform the tree traversal re-using the generate tree topology. An example input and output \ours generated by GPT-4 for emotion classification is shown in Figure \ref{fig:tree}.
 
\begin{table*}[h]
\centering
\scalebox{0.83}{
\begin{tabular}{ll|cccccc}
\toprule
\multirow{2}{*}{Model} & \multirow{2}{*}{Method} & \multicolumn{6}{c}{Performance (Num. Classes)} \\ 
\cmidrule{3-8} 
 && cola (2) & emotion (6) & amazon (5) & hsd (2) & sbic (3) & \textbf{Average} \\ 
\midrule
\multirow{4}{*}{RoBERTa} & Multi-class Prompting & 65.87 & 49.2 & 33.31 & 67.78 & 52.99 &  53.83 \\ 
 & Human-Generated Tree & \textbf{69.03} & 22.20 & 26.88 & 64.85 & \textbf{58.37} & 48.27 \\
 & \ours w/ GPT-3.5 & 56.66 & 35.1 & 33.46 & 67.36 & 38.25 & 46.17 \\ 
 & \ours w/ GPT-4 & 68.94 & \textbf{54.5} & \textbf{38.88} & \textbf{70.92} & 55.95 & \textbf{57.65} \\ 
\midrule
\multirow{4}{*}{DeBERTa} & Multi-class Prompting & 53.50 & 51.93 & 37.01 & 67.78 & 59.08  & 53.86 \\
 & Human-Generated Tree & \textbf{69.22} & 32.15 & 33.00  & \textbf{72.18} & 55.02  & 52.31 \\
 & \ours w/ GPT-3.5 & 49.66 & 39.00 & 36.18 & 70.29 & 52.49 & 49.52 \\
 & \ours w/ GPT-4 & 68.36 & \textbf{55.4} & \textbf{40.2} & 70.08 & \textbf{59.68} & \textbf{58.74} \\
\midrule
None & Model-free \ours w/o Tree & 69.13 & 40.55 & 25.76 & 59.62 & 37.63  & 46.54 \\
\bottomrule
\end{tabular}}
\caption{Zero-shot performance of different human and LLM-generated classification schemes. The GPT-4 generated decision trees consistently exhibit significant improvement. For model-free \ours, generated code can be executed on the entire validation set in 2 seconds and notably surpasses the random baseline, with cola notably matching the state-of-the-art performance. Results on SST2 benchmark is outlined in Appendix \ref{appendix-sst2}.}
\label{table:nlu-tree}
\end{table*}

 We compare \ours against two baseline methods. Our first baseline uses  the zero-shot classification method proposed in \cite{ge2023entailment} (``multi-class prompting''). This method uses the same entailment models but makes the prediction without the tree structure. Our second baseline asks a human expert to design a decision tree for each task also based on the SST-2 example. The results shown in Table \ref{table:nlu-tree} show that \ours generated by GPT-4 outperforms multi-class prompting and human-generated tree baselines on most datasets.


\textbf{Model-free NLEP}.  We also tried using the task-general prompt shown in \ref{appendix:p4t26} to generate NLEPs that directly use programs to solve these tasks.  These programs do not need any  neural models and are hence very efficient (e.g., finishing the entire validation set in about 2 seconds on CPUs). The results can be found in Table \ref{table:nlu-tree} (``Model-free \ours''). While not achieving the performance of entailment-based methods, the generated \ours significantly outperforms random baselines, suggesting that this may be a viable approach for quickly extracting simple and interpretable classifiers from LLMs.

\section{Discussion}
\label{sec-discussion}
\begin{table*}[h]
\centering
\scalebox{0.82}{
\begin{tabular}{l|ccc|c|cc|cc}
\toprule
\multicolumn{1}{c|}{\multirow{3}{*}{Tasks / Methods}} & \multicolumn{3}{c|}{CodeLlama7b}                         & NLEP-CodeLlama7b & \multicolumn{2}{c|}{CodeLlama13b} & \multicolumn{2}{c}{Claude2}   \\ 
\multicolumn{1}{c|}{}                                 & \multicolumn{1}{c}{\textcolor{gray}{Task-Specific}} & \multicolumn{2}{c|}{Task-General} & Zero-Shot                     & \multicolumn{2}{c|}{Task-General}               & \multicolumn{2}{c}{Task-General} \\ 
\multicolumn{1}{c|}{}                                 & \multicolumn{1}{c}{\textcolor{gray}{PoT}}           & PoT             & NLEP            & NLEP                          & PoT                    & NLEP                   & PoT             & NLEP           \\ \midrule
Tracking Shuffled Objects                         & \multicolumn{1}{c|}{\textcolor{gray}{95.6}}          & 21.2            & 30.0            & 84.4                          & 23.2                   & 23.2                   & 93.6            & 96.0           \\
Dyck Language                                         & \multicolumn{1}{c|}{\textcolor{gray}{15.2}}          & 0.8             & 0.8             & 1.2                           & 2.0                    & 2.8                    & 33.6            & 60.4           \\
Word Sorting                                          & \multicolumn{1}{c|}{\textcolor{gray}{78.0}}          & 98.0            & 93.2            & 98.4                          & 95.6                   & 97.6                   & 99.6            & 99.6           \\
Chinese Remainder Theorem                             & \multicolumn{1}{c|}{\textcolor{gray}{100.0}}         & 0.0             & 18.8            & 0.0                           & 1.2                    & 27.2                   & 40.4            & 60.4           \\
Scheduling Meeting                                    & \multicolumn{1}{c|}{\textcolor{gray}{32.0}}          & 4.0             & 24.8            & 34.4                          & 5.6                    & 8.8                    &      10.4$^\dag$           &     24.0$^\dag$           \\
\bottomrule
\end{tabular}}
\caption{Performance on five reasoning tasks adopted by \citet{cai2023large}: (a) Prompting CodeLlama-7b-Instruct \citep{rozière2023code} with task-specific and task-general demonstrations. (b) We train CodeLlama-7b \citep{rozière2023code} with out-of-domain \ours-format examples and report the zero-shot performance. (c) Prompting CodeLlama-13b-Instruct \citep{rozière2023code} with task-general demonstrations. (d) Prompting Claude-2 with task-general demonstrations by API. The demonstration examples remain consistent with these outlined in Table \ref{table:symbolic-reasoning}. $^\dag$ indicates human evaluation. 
}
\label{table:more-model-reasoning}
\end{table*}
\textbf{Execution failures and retries.} While the task-general PoT and \ours lack programming demonstrations for the target task, GPT-4 in general is able to generate bug-free programs as presented in Appendix \ref{appendix-3-additional-symbolic} Table \ref{table:execution-error}. Notably, both PoT and \ours obtain execution error rate of 0 on Tracking Shuffled Objects (7) and Word Sort tasks. 
One advantage of the program synthesis approaches such as PoT and NLEP is that non-executable programs can be identified and filtered. This gives LLMs the chance to ``self-correct" and generate new answers, and we take advantage of this in our math and symbolic reasoning tasks by generating up to three programs if there is an execution failure on certain benchmarks. (For fair comparison we apply this reattempting scheme to PoT as well). We ablate on this mechanism in Appendix \ref{appendix-3-additional-symbolic}, 
Tables \ref{table:symbolic-reasoning-details}, \ref{table:game24-details} and \ref{table:execution-error}.
Besides effectively reducing the execution error as presented in Table \ref{table:execution-error}, these retries greatly enhance the reasoning accuracy. In particular, 12\% and 15.6\% improvement is observed on the Chinese Remainder Theorem and the Scheduling Meeting tasks in Table \ref{table:symbolic-reasoning-details}(b). In this work we only experiment extra retries with larger temperatures for diverse sampling and leave more advanced ``self-correction" algorithms (e.g., those that make use of error messages \citep{cai2023large, hu2023code}) for future work. 


\textbf{Different foundation LLMs for \ours.} The performance of task-general PoT and \ours with more foundation models is reported in Table \ref{table:more-model-reasoning}. We provide detailed analysis in Appendix \ref{appendix-3-additional-symbolic}. Overall, \ours demonstates superior performance compared to task-general PoT, particularly excelling in complex tasks such as the Chinese Remainder Theorem. We note the trend of performance decline on the reasoning tasks as the abilities of underling LLMs decreased from GPT-4, Claude-2, GPT-3.5-Turbo to CodeLlama \citep{rozière2023code}. However, this observation is not limited to \ours, it also applies to other prevalent reasoning approaches like PoT. For example, on the Dyck Language task, GPT-3.5-Turbo only achieves 7.2\% (4.0\%) accuracy with \ours (PoT) prompting, while GPT-4 improves these figures to 91.6\% (66.4\%). It indicates that strong programming ability of underlying LLMs is vital to generate accurate responses and attain satisfactory performance. 
Surprisingly, zero-shot CodeLlama-7b \citep{rozière2023code} trained using \ours-style data (without in-domain examples) demonstrates superiority on Tracking Shuffled Objects (7) benchmark over \ours prompted GPT-3.5 and Word Sorting benchmark over task-general CoT prompted GPT-3.5, even with significantly fewer parameters.
It shows the potential for effective training of compact large language models, enabling them to achieve performance levels comparable to those of extremely large models.

\section{Related Work}
\noindent \textbf{Large language models for reasoning.}
State-of-the-art LLMs \citep{chatgpt2023,openai2023gpt4,touvron2023llama,zeng2022glm} have shown very strong performance on complicated reasoning tasks, including commonsense \citep{geva2021strategyqa}, math \citep{cobbe2021training}, symbolic reasoning \citep{suzgun2022challenging}, and programming \citep{austin2021program,chen2021codex}.  Tackling such tasks with LLMs often requires prompting them with demonstrations that elicit their reasoning capabilities. \cite{wei2022chain} proposed chain-of-thought prompting technique that encourages language to generate answers step-by-step. \cite{wang2022self} found that self-consistency can further improve the performance of chain of thoughts reasoning ability. \cite{kojima2023large} discovered that LLMs can perform reasoning without any demonstrations through adding the incantation
``\texttt{Let’s think step-by-step}".
Tree of thoughts \citep{yao2023tree} and graph of thoughts \citep{yao2023beyond,besta2023graph} were proposed to tackle tasks that require more complicated reasoning processes. These improved reasoning methods apply chain of thoughts as the atomic reasoning step but organize reasoning ``chains'' through more advanced mechanisms. 

\noindent \textbf{Programs and tools.} Previous studies have found that some limitations of LLMs can be overcome by combining program synthesis techniques with prompt-based learning.
Program of thoughts \citep{chen2022program} and program aided language model \cite{gao2023pal} both translate mathematical questions to equations and use the python interpreter to ensure the correctness of the calculations. 
Another line of related work for enabling LLMs to use tools is through interleaving API calls during LLM generation \citep{qin2023tool,liang2023taskmatrixai,mialon2023augmented,tang2023toolalpaca}.  APIs can aid many tasks that are challenging for LLMs by providing tailored tools (e.g., calculators, search) that can solve specific tasks. Toolformer \citep{schick2023toolformer} addresses reasoning tasks by using predefined tools, and LLMs as tool makers (LATM) can implement functions solving a class of tasks based on few-shot examples \citep{cai2023large}. With these solutions, the correctness of a prediction can be ensured if correct API is called and correct inputs are selected. Existing works on combining program synthesis and tool usage with LLMs generally rely on task-specific prompts, in contrast to the more task-general prompt explored in the present work.

\section{Conclusion}
This work describes natural language embedded programs (NLEP), which flexibly combine natural language reasoning with program synthesis within prompt-based learning to tackle a variety of tasks. Our experiments demonstrate that NLEPs expand the scope of  applications that can be addressed by program synthesis by more closely incorporating natural language during code generation. 

\section*{Acknowledgement}
This research was supported by the Center for Perceptual and Interactive Intelligence (CPII) Ltd under the Innovation and Technology Commission’s InnoHK Scheme.

\section*{Limitation} We found that the \ours prompts are not suitable for generating long-form natural language responses. Experimental results on VicunaQA show that most responses generated by \ours prompting have fewer tokens than responses obtained from usual LLM generation. This feature is expected, because most naturally-occurring programs (on which the LLMs were pretrained) do not contain large chunks of natural language. Future work could consider incorporating (potentially synthetically generated) programs with  longer-form natural language within the pretraining set to enable the application of NLEP to more involved NLG tasks. In the context of prompt-based approaches without parameter adjustments, the design of prompts can affect performance. While we aimed to mitigate this issue by expanding the scope of evaluation benchmarks and minimizing random factors, we did not engage in extensive prompt variation analysis. Further exploration into the impact of prompt variations remains an avenue for enhancing the robustness of reasoning approaches.

\section*{Ethical Statement}
This work intends to design a accurate and interpretable reasoning framework for language that entails more transparent and responsible LLM applications. However, the program generation method is more capable to handle different tasks in both areas of natural and program languages, infecting both humans and computing systems. As a result, we believe program generation models need stronger alignment and careful management.


\bibliography{anthology,custom}

\appendix
\clearpage


\section{Additional Examples}
\begin{figure*}[h]
\centering
\includegraphics[width=\textwidth]{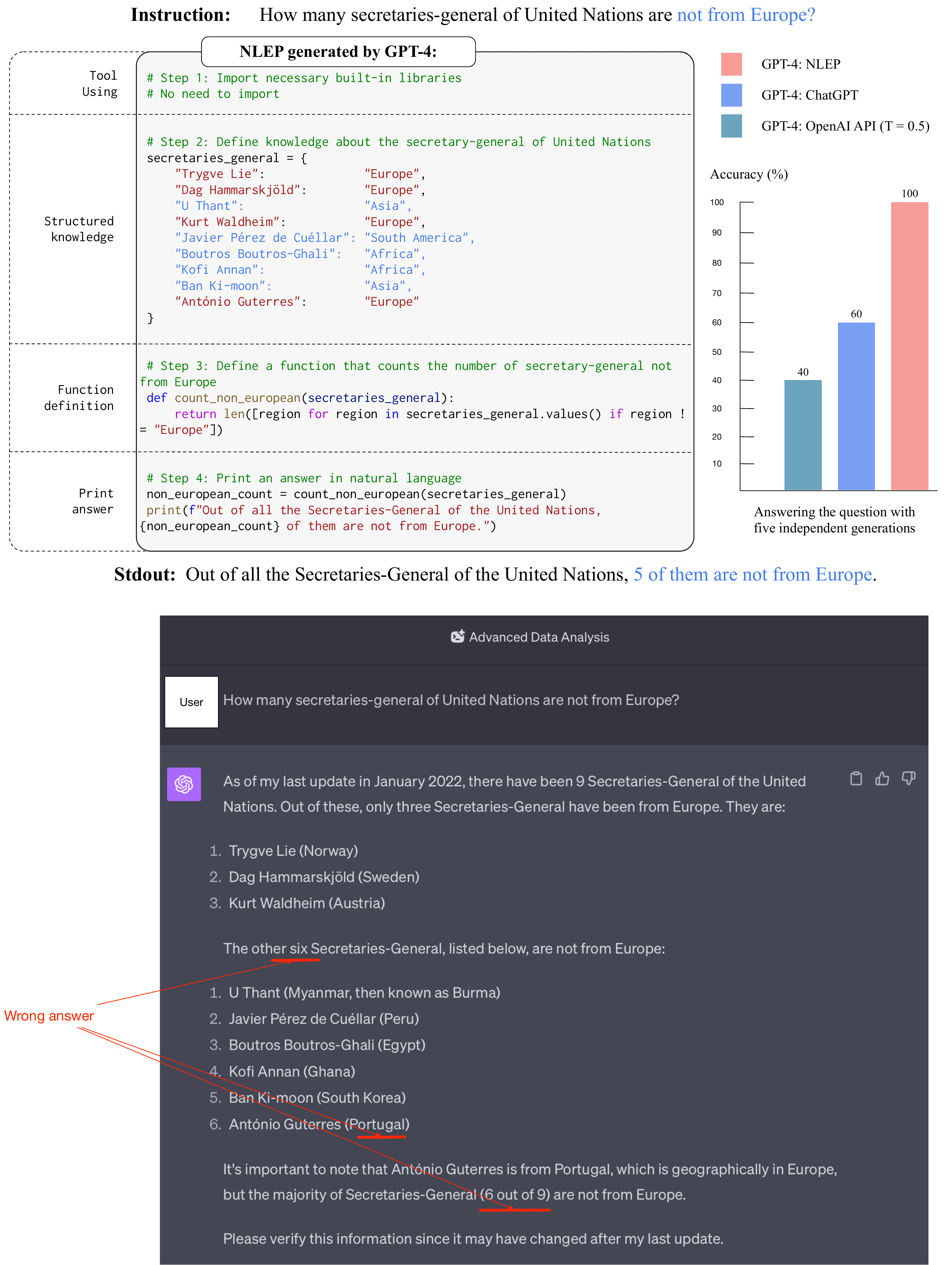}
\caption{\ours answering a question which requires numeric reasoning of structured knowledge. ChatGPT-4 code interpreter (currently the advanced data analysis option) constantly prefers to answer this question with plain natural language.
}
\label{fig:figure1}
\end{figure*}

\begin{figure*}[h]
\centering
\includegraphics[width=.9\textwidth]{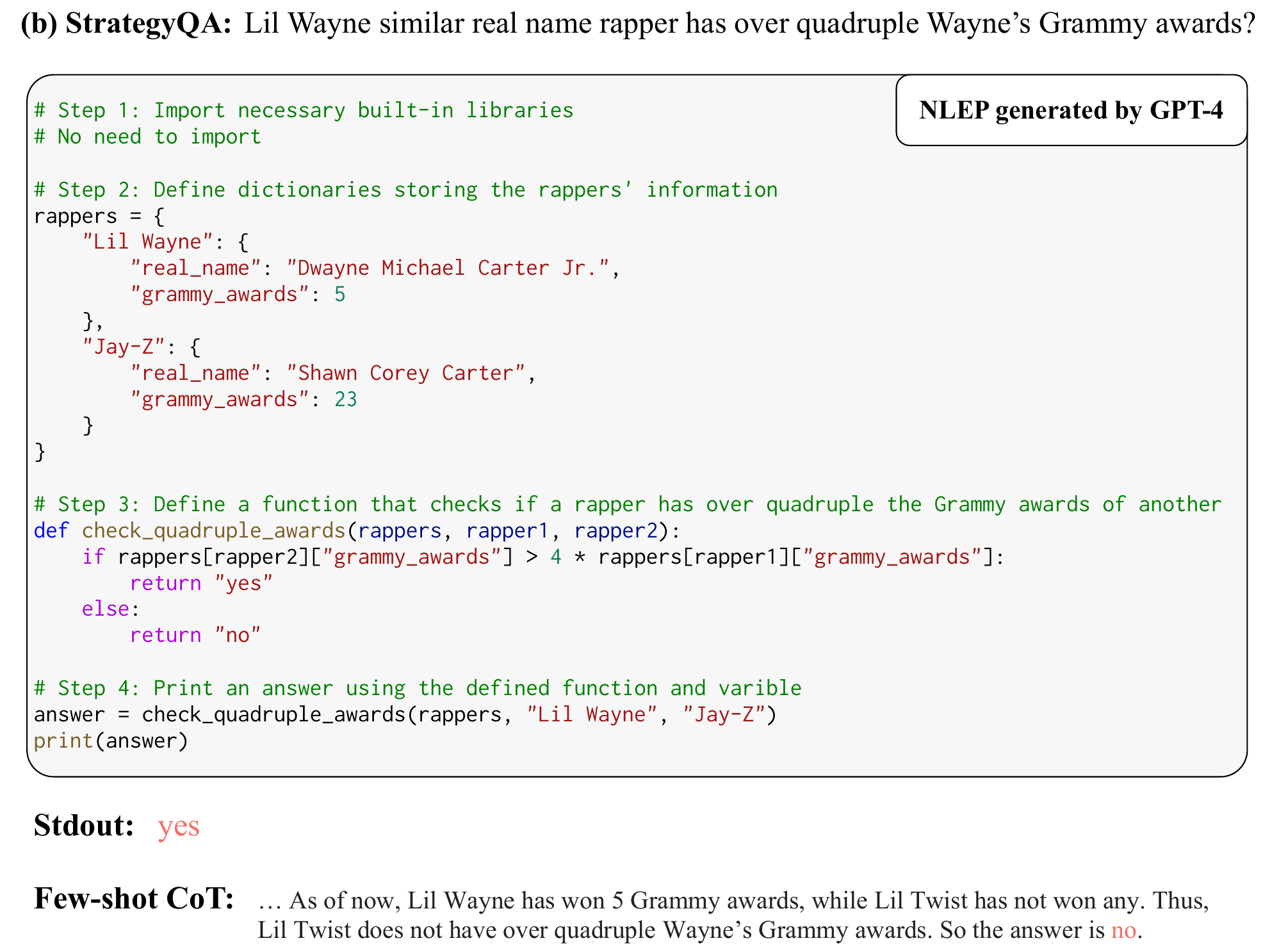}
\caption{\ours generated for solving StrategyQA problem. The instruction is \textit{``Answer the question with yes or no.''}
}
\label{fig:sqa-example}
\end{figure*}
A detailed comparison between NLEP and ChatGPT-4 is shown in Figure \ref{fig:figure1}. An example of generated \ours for StrategyQA problem is outlined in Figure \ref{fig:sqa-example}.

\section{Additional Results and Analysis on Math and Symbolic Reasoning}
\label{appendix-3-additional-symbolic}

We present the detailed experimental results of math and symbolic reasoning tasks in Tables \ref{table:symbolic-reasoning-details} to \ref{table:game24-details}, with execution failure analysis in Table \ref{table:execution-error}. The significance test analysis is outlined in Table \ref{table:significance-test}.

\textbf{GPT Results}. We report the results of Table \ref{table:symbolic-reasoning} with more details in Table \ref{table:symbolic-reasoning-details}. The effect of extra retries described in Section \ref{sec-discussion} is highlighted with ($\rightarrow$). The detailed experimental results on the Game of 24 benchmark is listed in Table \ref{table:game24-details}. 

\begin{table*}[h]
\centering
\scalebox{0.83}{
\begin{tabular}{l|cccccc}
\toprule
                & \begin{tabular}[c]{@{}c@{}}Tracking Shuffled \\ Objects (7)\end{tabular} & \begin{tabular}[c]{@{}c@{}}Dyck \\Language\end{tabular} & \begin{tabular}[c]{@{}c@{}}Word\\ Sorting\end{tabular}  & \begin{tabular}[c]{@{}c@{}}Chinese \\Remainder Theorem\end{tabular} & \begin{tabular}[c]{@{}c@{}}Scheduling\\Meeting\end{tabular}  & GSM-Hard \\
                \midrule

\multicolumn{7}{c}{\textcolor{gray}{(a) Task-Specific Prompting: GPT-4}}                                                                                                                                                                                     \\ \midrule
\textcolor{gray}{CoT} & \textcolor{gray}{\textbf{\underline{100.0}}} & \textcolor{gray}{63.6$^\dagger$}          & \textcolor{gray}{90.9$^\dagger$}        & \textcolor{gray}{0.0$^\dagger$}                                                                 & \textcolor{gray}{55.6$^\dagger$}               & \textcolor{gray}{57.4}    \\
\textcolor{gray}{PoT}             & \textcolor{gray}{\textbf{\underline{100.0}}}                                                                    & \textcolor{gray}{60.8}          & \textcolor{gray}{\textbf{\underline{100.0}}}        & \textcolor{gray}{\textbf{\underline{100.0}}}                                                               & \textcolor{gray}{75.2}               & \textcolor{gray}{\textbf{\underline{74.1}}}    \\
\textcolor{gray}{\textcolor{gray}{LATM}}            & \textcolor{gray}{\textbf{\underline{100.0}}}                                                                    & \textcolor{gray}{\underline{87.5}$^\dagger$}          & \textcolor{gray}{99.1$^\dagger$}        & \textcolor{gray}{\textbf{\underline{100.0}}$^\dagger$}                                                              & \textcolor{gray}{\textbf{\underline{100.0}}$^\dagger$}              & -        \\
\midrule
\multicolumn{7}{c}{(b) Task-General Prompting: GPT-4} \\ \midrule
CoT             & 81.2                                                                     & 39.6          & 84.4         & 0.0                                                                 & 82.8               & 54.9    \\
PoT             & 98.4                                                                     & 66.4          & \underline{99.6}         & 76.4 ($\rightarrow$84.4)                                                                & \underline{84.4} ($\rightarrow$85.2)             
& \underline{69.3}    \\
\ours (Ours) & \textbf{\underline{100.0}}                                                                    & \textbf{\underline{91.6}}          & \underline{99.6}         & \underline{85.2 ($\rightarrow$97.2)}                                                                & 77.6 \underline{($\rightarrow$93.2)}               
& 67.7   \\  \midrule 
\multicolumn{7}{c}{\textcolor{gray}{(c) Task-Specific Prompting: GPT-3.5-Turbo}} \\ \midrule
\textcolor{gray}{CoT}             & \textcolor{gray}{\underline{68.0}}& \textcolor{gray}{20.4$^\dagger$}& \textcolor{gray}{59.2$^\dagger$}& \textcolor{gray}{0.0$^\dagger$}& \textcolor{gray}{18.9$^\dagger$}& \textcolor{gray}{45.0}\\
\textcolor{gray}{PoT}             & \textcolor{gray}{6.8}& \textcolor{gray}{\underline{28.4}}& \textcolor{gray}{\textbf{\underline{100.0}}}& \textcolor{gray}{\textbf{\underline{100.0}}}& \textcolor{gray}{\underline{33.6}}& \textcolor{gray}{\underline{63.4}} \\\midrule 
\multicolumn{7}{c}{(d) Task-General Prompting: GPT-3.5-Turbo}                                                                                                                                                                                     \\ \midrule
CoT             & 51.2& \underline{38.0}& 75.2& 0.0 & 39.6& 42.8 \\
PoT             & \underline{88.4}& 4.0& \textbf{\underline{100.0}}& 58.4 ($\rightarrow$72.4)& 46.8 ($\rightarrow$49.2)& 39.0 ($\rightarrow$52.2)  \\
\ours (Ours)     & 74.4& 7.2&99.6 & \underline{94.8 ($\rightarrow$96.4)} & \underline{75.2 ($\rightarrow$85.6)}& \underline{50.9 ($\rightarrow$54.1)} \\ 
\bottomrule
\end{tabular}}
\caption{Performance on six reasoning tasks. $^\dagger$ stands for results from LATM \citep{cai2023large}. Results with $^\dagger$ and of LATM are reported on the last 240 instances with the first 10 instances as training and validation sets for tool making according to LATM's design. LATM is not appropriate for GSM-Hard benchmark as it is hard to derive a generally applicable tool function for all test cases. We mainly experiment LATM with GPT-4 as the tool maker since \cite{cai2023large} found that GPT-3.5 fails in all 5 trials on hard tasks like Tracking Shuffled Objects (5). If the generated tool is not general enough or only suitable for training samples, the tool using phase will fail.
We perform experiments using 
GPT-4 and GPT-3.5-Turbo with a temperature of 0 for all settings except PoT and \ours on GSM-Hard in (b) which use a temperature of 0.5 to increase the sampling diversity. Since task-general PoT and \ours lack task-specific programming instruction, they may generate non-executable Python programs. We select some settings and give each instance failed at execution up to three additional trials with temperature=0.4 to diversify the possible outputs. No label leakage is involved in this process as only the success of execution is used as a judgement. We report the results with these extra retries on execution failures in ($\rightarrow$). The highest score among each sub-table (a), (b), (c) and (d) is \underline{underlined} and the best result for each task is in \textbf{bold}.}
\label{table:symbolic-reasoning-details}
\end{table*}

\begin{table*}[h]
\centering
\scalebox{0.83}{
\begin{tabular}{l|cccccc|cc}
\toprule
\multirow{2}{*}{setting} & \multicolumn{6}{c|}{Task-Specific}                                                                                                                                                                                                                 & \multicolumn{2}{c}{Task-General} \\
                         & IO  & CoT & \begin{tabular}[c]{@{}c@{}}IO\\ (best of 100)\end{tabular} & \begin{tabular}[c]{@{}c@{}}CoT\\ (best of 100)\end{tabular} & \begin{tabular}[c]{@{}c@{}}ToT \\ (b=1)\end{tabular} & \begin{tabular}[c]{@{}c@{}}ToT \\ (b=5)\end{tabular} & PoT         & \ours (ours)         \\ \midrule
Game of 24               & 7.3$^\dagger$ & 4.0$^\dagger$ & 33.0$^\dagger$                                                       & 49.0$^\dagger$                                                        & 45.0$^\dagger$                                                 & \textbf{74.0$^\dagger$}                                                 & 
52 ($\rightarrow$52)
& 63 ($\rightarrow$66)                  \\ \bottomrule
\end{tabular}}
\caption{Performance on Game of 24 benchmark. $^\dagger$ stands for results from \cite{yao2023tree}. 
}
\label{table:game24-details}
\end{table*}

\textbf{Significance Test}. We report the significance test in Table \ref{table:significance-test} for task-general PoT and \ours prompted GPT-4 results. Due to API cost hamper, we run significance test using bootstrap test: \url{https://github.com/neubig/util-scripts/blob/master/paired-bootstrap.py}. The hyperparameters are configured to the default setting: \texttt{num\_samples=10000, sample\_ratio=0.5} Besides, to make the results more reliable and reproducible, we performed the major experiments using a temperature of 0 as reported in Table \ref{table:symbolic-reasoning-details}, and assessed \ours across a diverse range of tasks to show its general capabilities.

\begin{table*}[h]
\centering
\scalebox{0.8}{
\begin{tabular}{l|cc|ccl}
\toprule
\multirow{2}{*}{Tasks/Evaluation} & \multicolumn{2}{c|}{Accuracy} & \multicolumn{3}{c}{Significance Test}                                                                                \\
                                  & PoT           & NLEP          & NLEP win/lose/tie & 95\% Confidence Interval                                                      & P-value          \\ \midrule
Tracking Shuffled Objects (7)     & 98.4          & 100.0         & 86.8/0.0/13.2     & \begin{tabular}[c]{@{}l@{}}{[}1.000,1.000{]}\\ {[}0.960,1.000{]}\end{tabular} & 0.132            \\
Dyck Languages                    & 66.4          & 91.6          & 100.0/0.0/0.0     & \begin{tabular}[c]{@{}l@{}}{[}0.864,0.960{]}\\ {[}0.576,0.744{]}\end{tabular} & 0.000            \\
Word Sorting                      & 99.6          & 99.6          & 0.0/0.0/100.0     & \begin{tabular}[c]{@{}l@{}}{[}0.984,1.000{]}\\ {[}0.984,1.000{]}\end{tabular} & /                \\
Chinese Remainder Theorem         & 84.4          & 97.2          & 100.0/0.0/0.0     & \begin{tabular}[c]{@{}l@{}}{[}0.944,1.000{]}\\ {[}0.776,0.904{]}\end{tabular} & 0.000            \\
Scheduling Meeting                & 85.2          & 93.2          & 98.5/0.7/0.8      & \begin{tabular}[c]{@{}l@{}}{[}0.888,1.976{]}\\ {[}0.784,0.912{]}\end{tabular} & 0.015            \\
GSM-Hard                          & 69.3          & 67.7          & 0.92/88.8/20.0    & \begin{tabular}[c]{@{}l@{}}{[}0.642,0.712{]}\\ {[}0.657,0.728{]}\end{tabular} & PoT better 0.112 \\ \bottomrule
\end{tabular}}
\caption{We report the significance test results for task-general PoT and \ours prompted GPT-4 in Table \ref{table:symbolic-reasoning}. Due to API cost hamper, we run significance test using bootstrap test: \url{https://github.com/neubig/util-scripts/blob/master/paired-bootstrap.py}. The hyperparameters are configured to the default setting: \texttt{num\_samples=10000, sample\_ratio=0.5}.}
\label{table:significance-test}
\end{table*}

\begin{table*}[h]
\centering
\renewcommand{\arraystretch}{1.2}
\scalebox{0.82}{
\begin{tabular}{l|ccc|ccc}
\toprule
\multirow{3}{*}{}          & \multicolumn{3}{c|}{GPT-4}                                              & \multicolumn{3}{c}{GPT-3.5-Turbo}                                      \\ \cline{2-7} 
                           & \multicolumn{1}{c|}{Task-Specific} & \multicolumn{2}{c|}{Task-General} & \multicolumn{1}{c|}{Task-Specific} & \multicolumn{2}{c}{Task-General} \\
                           & \multicolumn{1}{c|}{PoT}           & PoT              & \ours            & \multicolumn{1}{c|}{PoT}           & PoT              & \ours           \\ \midrule
Track Shuffled Objects (7) & \multicolumn{1}{c|}{0}             & 0                & 0               & \multicolumn{1}{c|}{233}           & 26               & 24             \\
Dyck Language              & \multicolumn{1}{c|}{16}            & 24               & 10              & \multicolumn{1}{c|}{32}            & 81               & 26             \\
Word Sort                  & \multicolumn{1}{c|}{0}             & 0                & 0               & \multicolumn{1}{c|}{0}             & 0                & 0              \\
Chinese Remainder Theorem  & \multicolumn{1}{c|}{0}             & 32 ($\rightarrow$0)           & 37 ($\rightarrow$6)          & \multicolumn{1}{c|}{0}             & 46 ($\rightarrow$7)              & 4 ($\rightarrow$0)          \\
Scheduling Meeting         & \multicolumn{1}{c|}{0}             & 3 ($\rightarrow$0)            & 43 ($\rightarrow$0)          & \multicolumn{1}{c|}{2}             & 15 ($\rightarrow$2)           & 36 ($\rightarrow$0)         \\
GSM-Hard                   & \multicolumn{1}{c|}{17}            & 6                & 8               & \multicolumn{1}{c|}{31}            & 464 ($\rightarrow$145)        & 95 ($\rightarrow$13)        \\ \bottomrule
\end{tabular}}
\caption{Execution failure statistics on six math and symbolic reasoning tasks. Results with extra reties are reported in ($\rightarrow$). For task-specific PoT, we report the execution error statistics with \texttt{None} as the return value of \texttt{safe\_execute()} function following the {source code} of \cite{chen2022program}: \protect\url{https://github.com/wenhuchen/Program-of-Thoughts/blob/main/tool.py}. It includes instances where the generated programs do not contain the required line of code: \texttt{ans=\{desired\_result\}}, which are explicitly required in the instruction and given in the prompt demonstration. Under this scenario, we cannot capture the execution results of task-specific PoT.}
\label{table:execution-error}
\end{table*}

\textbf{Results with Different Fundation Models.} To investigate the effect of \ours on different large language models, we report the results with Claude-2\footnote{\url{https://www.anthropic.com/api}} and CodeLlama \citep{rozière2023code} in Table \ref{table:more-model-reasoning}. Following the guidance of the instruction-following models\footnote{\url{https://github.com/facebookresearch/codellama}}, we employ a chat session to include task-specific and task-general prompts as previous turns by interleaving the ``user" and ``assistant" messages with a system message ``Provide answers in Python" at the beginning for CodeLlama-7b-Instruct and CodeLlama-13b-Instruct. Hence,
we only treat bug-free Python programs that have the desired results after execution as correct answers, regardless of natural language outputs since we explicitly prompt CodeLlama to generate the answer in Python. Unlike the prominent performance of GPT-4, the positive impact of \ours with CodeLlama is diminished due to the much smaller model size and greatly reduced programming capability. Although \ours prompting outperforms the task-general PoT by a large margin on Chinese Remainder Theorem and Scheduling Meeting benchmarks, a non-negligible performance gap is observed between \ours and task-specific PoT on most tasks. We notice a decline or performance on Scheduling Meeting bechmark of 13b model over 7b. Interestingly, we found that CodeLlama-13b prefers to use tools (built-in library) more for solving Scheduling Meeting problems while CodeLlama-7b tends to do it directly “by hand”. However, the tool use of a less-proficient model may result in more possible execution errors. For instance, 113 over 125 execution errors (total 250 test cases) of CodeLlama-13b is \texttt{\color{blue}{Error: `datetime.time' object has no attribute `overlaps'}} while CodeLlama-7b only has 3 execution errors.
The performance with Claude-2 is more prominent, with \ours demonstrates superior performance on Dyck Language, Scheduling Meeting and Chinese Remainder Theorem benchmarks. 


To further investigate the benefits of \ours, we fine-tune a CodeLlama-7b \citep{rozière2023code} model using \ours-style instances, resulting in a variant that we term \ours-CodeLlama. Note that our training corpus does not include specific evaluation tasks.
During the evaluation phase, we adopt zero-shot prompting strategy, where the model is provided with only test instances without in-context demonstrations. As presented in Table \ref{table:more-model-reasoning}(b),
zero-shot \ours-CodeLlama exhibits consistent performance improvement on 5 of 6 tasks. The only exception is the Chinese Remainder Theorem benchmark, which is notably more complex in nature. Remarkably, zero-shot \ours-CodeLlama demonstrates superior performance on Word Sorting benchmark when compared to task-general CoT prompted GPT-3.5-Turbo, and outperforms \ours prompted GPT-3.5-Turbo on Tracking Shuffled Objects (7) benchmark, despite a considerably lower parameter size.


\textbf{Limitation of GSM-Hard Benchmark.} We opted for GSM-Hard \citep{gao2023pal} to mitigate potential data contamination from GSM8K. The numbers in GSM-Hard are large, less likely to have been encountered by LLMs during training, and hence augmenting the task complexity. However, since GSM-Hard is automatically constructed by replacing small numbers in GSM-8K with large ones, there are peculiar cases where sometimes models may refuse to reply or automatically solve it. An example is as follows: the original question in GSM8K is ``\textit{Claire makes a 3 egg omelet every morning for breakfast. How many dozens of eggs will she eat in 4 weeks?}'' The number \texttt{3} is replaced with \texttt{6022727} and the corresponding question in GSM-Hard is ``\textit{Claire makes a 6022727 egg omelet every morning for breakfast. How many dozens of eggs will she eat in 4 weeks?}'' The output of task-general PoT aligns with the ground-truth, which is $6022727\times7\times4/12=14053029.667$, a decimal number. However, GPT-4 with \ours prompting regards the number of dozens should be an integer, and consequently returns the result as $6022727\times7\times4//12=14053029$. This gives additional challenges in evaluation, and we thus mainly evaluate the former five reasoning benchmarks using more models in Table \ref{table:more-model-reasoning}, which might serve as more reliable performance indicators.


\textbf{Execution Failure Analysis.} We present the execution failure statistics of code-based reasoning strategies in Table \ref{table:execution-error}. The effect of extra retries described in Section \ref{sec-discussion} is highlighted with ($\rightarrow$). Note that different from task-specific PoT with demonstrations showing how to return the desired outputs in Python program, e.g.,

\begin{lstlisting}[
    basicstyle=\scriptsize
]
# Python code, return ans
Alice = "striker"
Bob = "right winger"
Claire = "left winger"
Dave = "benchwarmer"
Eve = "goalkeeper"
Fred = "center midfielder"
Gertrude = "cheerleader"
Eve, Claire = Claire, Eve
Gertrude, Alice = Alice, Gertrude
Fred, Bob = Bob, Fred
Dave, Fred = Fred, Dave
Fred, Bob = Bob, Fred
Bob, Eve = Eve, Bob
Claire, Alice = Alice, Claire
ans = Gertrude
\end{lstlisting}

we need to design rules to extract the target answers from the execution results of task-general PoT and \ours since they are allowed to generate free-from outputs. For example, given the generated programs,

\begin{lstlisting}[
    basicstyle=\scriptsize,
    linewidth=\columnwidth,
    breaklines=true
]
# Step 1: Import necessary built-in libraries
# No need to import

# Step 2: Define necessary functions that generally solve this type of problem
def swap_positions(positions, swaps):
    for swap in swaps:
        positions[swap[0]], positions[swap[1]] = positions[swap[1]], positions[swap[0]]
    return positions

# Step 3: Define constant variables for the task
positions = {
    "Alice": "striker",
    "Bob": "right winger",
    "Claire": "left winger",
    "Dave": "benchwarmer",
    "Eve": "goalkeeper",
    "Fred": "center midfielder",
    "Gertrude": "cheerleader"
}

swaps = [
    ("Eve", "Claire"),
    ("Gertrude", "Alice"),
    ("Fred", "Bob"),
    ("Dave", "Fred"),
    ("Fred", "Bob"),
    ("Bob", "Eve"),
    ("Claire", "Alice")
]

# Step 4: Print an answer in natural language.
final_positions = swap_positions(positions, swaps)
print(f"At the end of the match, Gertrude is playing {final_positions['Gertrude']}.")
\end{lstlisting}

we need to extract ``striker", the target answer, from the execution results ``At the end of the match, Gertrude is playing striker.".

Although task-specific PoT explicitly instructs the model to generate a required line of code \texttt{ans=\{desired\_result\}} to capture the final result  by adding 
\begin{lstlisting}[
    basicstyle=\scriptsize,
    linewidth=\columnwidth,
    breaklines=true
]
# Python code, return ans
\end{lstlisting}
in the instruction and explicitly provides task-specific examples in the prompt, e.g.,

\begin{lstlisting}[
    basicstyle=\scriptsize,
    linewidth=\columnwidth,
    breaklines=true
]
# Python code, return ans
... ...
ans = Gertrude
\end{lstlisting}

we found that GPT models sometimes fail to accomplish the requirement. On Tracking Shuffled Objects (7) benchmark, only 17 programs out of 250 test cases generated by GPT-3.5 successfully contain the required line of code, i.e., \texttt{ans=\{desired\_result\}}, which explains the huge number of failures (233). In addition to already having task-specific programming examples, the inability to distinguish between the erroneous programs and lack of required line of code is another reason why we do not apply the error retries on task-specific PoT.

\textbf{\ours v.s. Task-general PoT} Compared to PoT, \ours offers structured and modularized Python blocks. This disentanglement of knowledge extraction from problem solving facilitates enhanced interpretability and controllability. 
In contrast, PoT may interwine different components, particularly blending knowledge and reasoning steps. An illustrative example is provided below: the question from GSM-Hard is \textit{Toula went to the bakery and bought various types of pastries. She bought 3 dozen donuts which cost 5804528 per dozen, 2 dozen mini cupcakes which cost 80 per dozen, and 6 dozen mini cheesecakes for 55 per dozen. How much was the total cost?} The generated Python block of \ours is
\begin{lstlisting}[
    basicstyle=\scriptsize,
    linewidth=\columnwidth,
    breaklines=true
]
# Step 1: Import necessary built-in libraries
# No need to import

# Step 2: Define necessary functions that generally solve this type of problem
def calculate_total_cost(items):
   total_cost = 0
   for item in items:
       total_cost += item['quantity'] * item['price']
   return total_cost

# Step 3: Define constant variables for the task
items = [
   {"name": "donuts", "quantity": 3, "price": 5804528},
   {"name": "mini cupcakes", "quantity": 2, "price": 80},
   {"name": "mini cheesecakes", "quantity": 6, "price": 55},
]

# Step 4: Print an answer in natural language.
total_cost = calculate_total_cost(items)
print(f"The total cost of the pastries that Toula bought from the bakery is ${total_cost}. This includes the cost of {items[0]['quantity']} dozen {items[0]['name']} at ${items[0]['price']} per dozen, {items[1]['quantity']} dozen {items[1]['name']} at ${items[1]['price']} per dozen, and {items[2]['quantity']} dozen {items[2]['name']} at ${items[2]['price']} per dozen.\nThe correct answer is ${total_cost}.")
\end{lstlisting}
The code returned by PoT is
\begin{lstlisting}[
    basicstyle=\scriptsize,
    linewidth=\columnwidth,
    breaklines=true
]
donuts = 3 * 5804528
cupcakes = 2 * 80
cheesecakes = 6 * 55

total_cost = donuts + cupcakes + cheesecakes
print(total_cost)
ans = Gertrude
\end{lstlisting}

\begin{table}[]
\centering
\scalebox{0.8}{
\begin{tabular}{l|ccccc}
\toprule
Num                    & cola & emotion & amazon & hsd & sbic \\
\midrule
Internal Node & 4    & 7       & 8      & 4   & 3    \\
Leaf Node              & 4    & 8       & 10     & 5   & 4    \\
Label                  & 2    & 6       & 5      & 2   & 3   \\ \bottomrule
\end{tabular}}
\caption{Statistics of NLEP generated decision trees for each benchmark in Table \ref{table:nlu-tree}.}
\label{table:node-num}
\end{table}

\section{Additional Results on Text Classification}
\label{appendix-sst2}
\textbf{Statistics of Decision Trees.} We list the statics of the generated decision tree by NLEP prompting in Table \ref{table:node-num}. Each internal decision node in the generated tree has two branches, “yes” and “no” , indicating whether it satisfies the requirement or not.

\noindent \textbf{Classification Results on SST2 Benchmark.} Since we use the trees derived from SST2 to prompt the LLM for the classification tasks, it would be inappropriate to use these examples for SST2 itself. For SST2, we thus use an automatically generated decision tree for the CoLA task to prompt GPT-4 to generate a new tree for SST2. As shown in Table \ref{table:sst2-tree}, the automatically generated tree matches the performance of the SST2 decision tree created by the authors.

\begin{table}[h]
\centering
\begin{tabular}{ll}
\toprule
\multicolumn{2}{c}{\textbf{SST2 Performance}}  \\
\midrule
\multicolumn{1}{l|}{Model-Free}        & 66.17 \\
\multicolumn{1}{l|}{RoBERTa-Manual}    & 83.03 \\
\multicolumn{1}{l|}{RoBERTa-Automatic} & 87.36 \\
\multicolumn{1}{l|}{DeBERTa-Manual}    & 84.06 \\
\multicolumn{1}{l|}{DeBERTa-Automatic} & 93.49 \\ \bottomrule
\end{tabular}
\caption{Performance of manually crafted vs. generated decision trees on SST2.}
\label{table:sst2-tree}
\end{table}

\clearpage

\section{Prompts for Task-General Strategies}
\label{appendix-1-prompt}

\subsection{Prompts for Table \ref{table:symbolic-reasoning} and \ref{table:strategyqa}}
\label{appendix:p4t26}
We list the prompts for the task-general chain-of-thought (CoT), our implementation of program-of-thoughts (PoT) and the proposed natural language embedded programs (\ours) strategies in the following code segments.
They share the same two \ours examples (one is for natural language reasoning and the other is for mathematical reasoning) but with different forms of intermediate reasoning steps (e.g., code, text etc.) to evaluate the average performance of different strategies. 

\textbf{Prompt for task-general chain-of-thought (CoT) in Table \ref{table:symbolic-reasoning} and \ref{table:strategyqa}.} The detailed intermediate natural language reasoning chains are generated by prompting GPT-4 given the input and target output.

\begin{lstlisting}[
    basicstyle=\scriptsize,
    linewidth=\columnwidth,
    breaklines=true
]
'''
Answer the problem based on the given instruction and input.

### Instruction: Identify the odd one out.
### Input: Twitter, Instagram, Telegram
### Answer:
Let's think step by step.
1. Start by understanding the task instruction. The task is to identify 
the odd one out from a given list.
2. Look at the input. The input consists of three items: Twitter, 
Instagram, and Telegram.
3. Identify what these items are. Twitter and Instagram are social media 
platforms, while Telegram is a messaging app.
4. Compare the items to find the odd one out. Twitter and Instagram are 
primarily used for sharing information, images, and videos. On the other 
hand, Telegram is mainly used for instant messaging and voice-over-IP 
service.
5. Determine the odd one out based on the comparison. In this case, 
Telegram is the odd one out because it serves a different primary 
function compared to Twitter and Instagram.
6. Formulate the target output. The target output should clearly state 
that Telegram is the odd one out and provide the reason why it is so. The 
reason being that Twitter and Instagram are social media platforms mainly 
for sharing information, images, and videos while Telegram is a cloud-
based instant messaging and voice-over-IP service.
The correct answer is Telegram.


### Instruction: Use the given data to calculate the median.
### Input: [2, 3, 7, 8, 10]
### Answer:
Let's think step by step.
1. Start by understanding the task, which is to calculate the median of a 
given data set. The median is the middle value in a sorted, ascending or 
descending, list of numbers.
2. Look at the given input, which is a list of numbers: [2, 3, 7, 8, 10]. 
3. Notice that the list is already sorted in ascending order. If it wasn'
t, the first step would be to sort it.
4. Understand that to find the median, we need to find the middle value. 
If the list has an odd number of observations, the median is the middle 
number. If the list has an even number of observations, the median is the 
average of the two middle numbers.
5. Count the number of values in the list. There are 5 values, which is 
an odd number, so the median will be the middle value.
6. Identify the middle value. Since there are 5 values, the middle value 
is the 3rd value when counting from either end.
7. Find the 3rd value in the list, which is 7.
8. Conclude that the median of the given data set is 7.
The correct answer is 7.
'''
\end{lstlisting}

\textbf{Prompt for task-general \ours in Table \ref{table:symbolic-reasoning} and \ref{table:strategyqa}.} The intermediate program reasoning chains are generated by prompting GPT-4 given the input and target output.

\begin{lstlisting}[
    basicstyle=\scriptsize,
    linewidth=\columnwidth,
    breaklines=true
]
Write a bug-free Python program that can generate the answer to the given instruction when correctly executed.

### Instruction: Identify the odd one out.
### Input: Twitter, Instagram, Telegram
### Python program:
```
# Step 1: Import necessary built-in libraries
from collections import OrderedDict

# Step 2: Define necessary functions that generally solve this type of problem
def find_odd_one_out(services, input_services):
    descriptions = [services[service] for service in input_services]
    for description in descriptions:
        if descriptions.count(description) == 1:
            return input_services[descriptions.index(description)]
    return None

# Step 3: Define constant variables for the task
services = OrderedDict([
    ("Twitter", "a social media platform mainly for sharing information, images and videos"),
    ("Instagram", "a social media platform mainly for sharing information, images and videos"),
    ("Telegram", "a cloud-based instant messaging and voice-over-IP service"),
])

input_services = ["Twitter", "Instagram", "Telegram"]

# Step 4: Print an answer in natural language.
odd_one_out = find_odd_one_out(services, input_services)
if odd_one_out:
    other_services = [service for service in input_services if service != odd_one_out]
    print(f"The odd one out is {odd_one_out}. {other_services[0]} and {other_services[1]} are {services[other_services[0]]} while {odd_one_out} is {services[odd_one_out]}.\nThe correct answer is {odd_one_out}.")
```

### Instruction: Use the given data to calculate the median.
### Input: [2, 3, 7, 8, 10]
### Python program:
```
# Step 1: Import necessary built-in libraries
# No need to import

# Step 2: Define necessary functions that generally solve this type of problem
def calculate_median(data):
    data.sort()
    length = len(data)
    if length % 2 == 0:
        return (data[length//2] + data[length//2 - 1]) / 2
    else:
        return data[length//2]

# Step 3: Define constant variables for the task
data = [2, 3, 7, 8, 10]

# Step 4: Print an answer in natural language.
median = calculate_median(data)
print(f"To find the median of a data set, we need to arrange the data in ascending order and then find the middle value. In this case, the given data is already arranged in ascending order. Since there are {len(data)} values in the data set, the median will be the middle value, which is the {len(data)//2 + 1}rd value. Hence, the median of the given data set is {median}.\nThe correct answer is {median}.")
```
\end{lstlisting}

\textbf{Prompt for our task-general program-of-thoughts (PoT) implementation in Table \ref{table:symbolic-reasoning} and \ref{table:strategyqa}.} The intermediate program reasoning chains are obtained from \ours. We keep the key implementation steps of Python programs but eliminate the step-by-step programming instructions and detailed natural language elaboration as an ablation.

\begin{lstlisting}[
    basicstyle=\scriptsize,
    linewidth=\columnwidth,
    breaklines=true
]
Write a bug-free Python program that can generate the answer to the given instruction when correctly executed.

### Instruction: Identify the odd one out.
### Input: Twitter, Instagram, Telegram
### Python program:
```
softwares = {
    'social media': ['twitter', 'instagram'],
    'communication': ['telegram']
}
for genre, apps in softwares.items():
    if len(apps) == 1:
        print(apps[0])
```

### Instruction: Use the given data to calculate the median.
### Input: [2, 3, 7, 8, 10]
### Python program:
```
data = [2, 3, 7, 8, 10]
data.sort()
length = len(data)
if length % 2 == 0:
    print((data[length//2] + data[length//2 - 1]) / 2)
else:
    print(data[length//2])
```
\end{lstlisting}

\subsection{Prompts for Table \ref{table:game24}}
The Game of 24 task is much more challenging and we replace the first example in Appendix \ref{appendix:p4t26} with a word sorting example
to elicit stronger reasoning ability.
\label{appendix:p4t3}

\textbf{Prompt for task-general \ours in Table \ref{table:game24}.} The intermediate program reasoning chains are generated by prompting GPT-4 given the input and target output.

\begin{lstlisting}[
    basicstyle=\scriptsize,
    linewidth=\columnwidth,
    breaklines=true
]
Write a bug-free Python program that can generate the answer to the given instruction when correctly executed.

### Instruction: Arrange the following words to make the longest possible word.
### Input: the, had, not, been
### Python program:
```
# Section 1: Define necessary functions and calculate intermediate variables
def longest_word(words):
    from itertools import permutations
    all_words = [''.join(p) for p in permutations(''.join(words))]
    all_words.sort(key=len, reverse=True)
    with open('english_words.txt') as word_file:  # Assuming you have a list of english words
        english_words = set(word.strip().lower() for word in word_file)
    for word in all_words:
        if word.lower() in english_words:
            return word
    return None

# Section 2: Define constant variables
words = ["the", "had", "not", "been"]

# Section 3: Insert variables in text outputs using f-strings.
longest = longest_word(words)
if longest:
    print(f"The longest word that can be made from the letters in the words \"{', '.join(words)}\" is \"{longest}\".")
```

### Instruction: Use the given data to calculate the median.
### Input: [2, 3, 7, 8, 10]
### Python program:
```
# Step 1: Import necessary built-in libraries
# No need to import

# Step 2: Define necessary functions that generally solve this type of problem
def calculate_median(data):
    data.sort()
    length = len(data)
    if length % 2 == 0:
        return (data[length//2] + data[length//2 - 1]) / 2
    else:
        return data[length//2]

# Step 3: Define constant variables for the task
data = [2, 3, 7, 8, 10]

# Step 4: Print an answer in natural language.
median = calculate_median(data)
print(f"To find the median of a data set, we need to arrange the data in ascending order and then find the middle value. In this case, the given data is already arranged in ascending order. Since there are {len(data)} values in the data set, the median will be the middle value, which is the {len(data)//2 + 1}rd value. Hence, the median of the given data set is {median}.")
```
\end{lstlisting}

\textbf{Prompt for our task-general program-of-thoughts (PoT) implementation in Table \ref{table:game24}.} The intermediate program reasoning chains are obtained from \ours. We keep the key implementation steps of Python programs but eliminate the step-by-step programming instructions and detailed natural language elaboration as an ablation.

\begin{lstlisting}[
    basicstyle=\scriptsize,
    linewidth=\columnwidth,
    breaklines=true
]
Write a bug-free Python program that can generate the answer to the given instruction when correctly executed.

### Instruction: Arrange the following words to make the longest possible word.
### Input: the, had, not, been
### Python program:
```
def longest_word(words):
    from itertools import permutations
    all_words = [''.join(p) for p in permutations(''.join(words))]
    all_words.sort(key=len, reverse=True)
    with open('english_words.txt') as word_file:  # Assuming you have a list of english words
        english_words = set(word.strip().lower() for word in word_file)
    for word in all_words:
        if word.lower() in english_words:
            return word
    return None

words = ["the", "had", "not", "been"]

longest = longest_word(words)
if longest:
    print(longest)
```

### Instruction: Use the given data to calculate the median.
### Input: [2, 3, 7, 8, 10]
### Python program:
```
data = [2, 3, 7, 8, 10]
data.sort()
length = len(data)
if length % 2 == 0:
    print((data[length//2] + data[length//2 - 1]) / 2)
else:
    print(data[length//2])
```
\end{lstlisting}

\subsection{Prompts for \ours in Table \ref{table:tqa} and Figure \ref{table:vicunaqa}}
\label{appendix:p4t7f5}
For experiments in TruthfulQA and VicunaQA, we added the following example into the \ours prompts shown in Appendix \ref{appendix:p4t26} to encourage generating more detailed responses:

\begin{lstlisting}[
    basicstyle=\scriptsize,
    linewidth=\columnwidth,
    breaklines=true
]
# Write a bug-free Python program that can generate the answer to the given instruction when correctly executed. Do not ask for user input. For reasoning tasks, define functions first and then define variables. For knowledge intensive tasks, define variables before defining functions. Do not define any variable that directly stores the final answer. If there is a knowledge definition step, use dictionaries to store both the knowledge and detailed explanation.

### Instruction: Discuss the causes of the Great Depression
### Input: None
### Python program:
```
# Step 1: Import necessary built-in libraries
# No need to import

# Step 2: Define dictionaries storing detailed knowledge about the grat depression
depression_name = "The Great Depression"
depression_period = "1929-1939"
depression_countries = "the United States and countries around the world"
depression_causes = {
    "Stock Market Crash of 1929": "In October of 1929, the stock market experienced a significant fall that wiped out millions of investors. This event is considered by many to be the initial trigger of the Great Depression.",
    "Overproduction": "During the 1920s, many industries produced more goods than consumers wanted or could afford. This ultimately led to a decline in demand for goods, causing job loss, lower wages, and business failure.",
    "High Tariffs and War Debts": "Protectionist trade policies in the form of high tariffs led to a decline in global trade, as other countries retaliated with tariffs of their own. Additionally, many countries were struggling to repay war debts, which led to economic instability.",
    "Bank Failures": "As demand for goods declined, many banks began to fail, causing a loss of confidence in the banking system. This led to a massive withdrawal of money from banks, causing even more banks to fail.",
    "Drought Conditions": "The Dust Bowl was a severe drought and dust storm that hit the Great Plains region of the United States in the 1930s. This had a significant impact on agriculture, causing many farmers to lose their land and livelihoods which worsened the effects of the depression."
}

# Step 3: Define necessary functions that generally solve this type of problem
# Do not need to define functions

# Step 4: Print the answer and explain in natural language by calling the information in the defined knowledge dictionary `depression_causes`
print(f"{depression_name} was a period of economic decline that lasted from {depression_period}, making it the longest-lasting depression in modern history. It affected not only {depression_countries}, causing substantial social and economic upheaval.\n")
print(f"There were several major causes of {depression_name}, which include:\n")

# List causes and explanations in `depression_causes` with a for-loop.
for i, (cause, description) in enumerate(depression_causes.items(), 1):
    print(f"{i}. {cause} - {description}\n")
print(f"Overall, {depression_name} was caused by a combination of factors, including economic, environmental, and political factors. Its impact was widespread, affecting millions of people around the world.")
```

### Instruction: Identify the odd one out.
### Input: Twitter, Instagram, Telegram
### Python program:
```
# Step 1: Import necessary built-in libraries
from collections import OrderedDict

# Step 2: Define dictionaries storing detailed knowledge about the main function of each application
services = {
    "Twitter": "a social media platform mainly for sharing information, images and videos",
    "Instagram": "a social media platform mainly for sharing information, images and videos",
    "Telegram": "a cloud-based instant messaging and voice-over-IP service",
}

# Step 3: Define a function that finds the different application
def find_odd_one_out(services, input_services):
    descriptions = [services[service] for service in input_services]
    for description in descriptions:
        if descriptions.count(description) == 1:
            return input_services[descriptions.index(description)]
    return None

# Step 4: Print the answer in natural language by calling the information stored in `services` and the defined function `find_odd_one_out`
input_services = ["Twitter", "Instagram", "Telegram"]
odd_one_out = find_odd_one_out(services, input_services)
if odd_one_out:
    other_services = [service for service in input_services if service != odd_one_out]
    print(f"The odd one out is {odd_one_out}. {other_services[0]} and {other_services[1]} are {services[other_services[0]]} while {odd_one_out} is {services[odd_one_out]}.")
```



### Instruction: Calculate the total surface area of a cube with a side length of 5 cm.
### Input:  None
### Python program:
```
# Step 1: Import necessary built-in libraries
# No need to import

# Step 2: Define a function that calculate the surface area of cubes
def calculate_surface_area(side_length):
    return 6 * (side_length ** 2)

# Step 3: Define dictionaries storing the cube information
cube = {
    "side_length": 5  # Side length of the cube
}

# Step 4: Print a step-by-step calculation answer in natural language using the defined function and varible
side_length = cube["side_length"]
surface_area = calculate_surface_area(side_length)
print(f"The surface area of a cube is found by calculating the area of one of its faces and multiplying it by six (since a cube has six faces). The area of a cube face is simply its side length squared.\n")
print(f"Thus for this particular cube:")
print(f"Surface Area = 6 x (Side Length)\^2")
print(f"             = 6 x ({side_length} cm)\^2")
print(f"             = 6 x {side_length**2} cm\^2")
print(f"             = {surface_area} cm\n")
print(f"The total surface area of this cube is {surface_area} square centimeters.")
```
```
\end{lstlisting}

\subsection{Prompts for Table \ref{table:nlu-tree} and \ref{table:sst2-tree}}
\label{appendix:p4t45}
We use the following prompt for the entailment-based \ours results in Table \ref{table:nlu-tree}. The model-free result uses the \ours prompt shown in \ref{appendix:p4t26}.

\begin{lstlisting}[
    basicstyle=\scriptsize,
    linewidth=\columnwidth,
    breaklines=true
]
"""Write a Python function that constructs a decision tree according to the given examples that can generate the correct label of the given classification task."""

### Available functions (shared for all tasks):

# Returns whether the hypothesis in entailed by the premise.
def entailment(hypothesis, premise, model, tokenizer):
    proposition = f'{hypothesis} is entailed by {premise}.'
    inputs = tokenizer(proposition,  return_tensors="pt", truncation=True, padding=True, max_length=128)
    outputs = model(**inputs)['logits'][0]
    ent_label = int(outputs[0] > outputs[2])
    if ent_label == 1:
        return 'yes'
    else:
        return 'no'

# Use the constructed decision tree to predict the label of the sentence.
def tree_predict(sentence, criterions, tree, model, tokenizer):
    node = tree['root']
    while node not in POSSIBLE_CLASSES:
        ent_label = entailment(criterions[node], sentence, model, tokenizer)
        node = tree[node][ent_label]
    return node

### Task: Movie review classification
### Possible classes: [positive, negative]
### Examples:
"""
- contains no wit, only labored gags
    - [The movie is wise|The movie is not wise|1], [the story is fun|the story is not boring|1], [the review is positive|the review is negative|1]
- that loves its characters and communicates something rather beautiful about human nature
    - [The characters are lovely|The characters are awful|0], [the script is touching|the script is dry|0], [the review is positive|the review is negative|0]
- on the worst revenge-of-the-nerds cliches the filmmakers could dredge up
    - [The movie is novel|The movie is mostly platitudes|1], [the review is negative|1]
- are more deeply thought through than in most right-thinking films
    - [The takeaway of the movie is profound|The idea of the movie is shallow|0], [the review is positive|the review is negative|0]
"""

### Define possible classes
POSSIBLE_CLASSES = ['positive', 'negative']

### Python program: 
```
def get_decision_tree(sentence, model, tokenizer):
    # Step 1: define criterions of the decision tree.
    criterions = [
        'This movie is interesting',
        'The movie has a good script',
        'The characters are awsome',
        'This movie is wise'
    ]

    # Step 2: define the Decision Tree for classification
    tree = {
        'root': 0,
        0: {'yes': 1, 'no': 3},
        1: {'yes': 'positive', 'no': 2},
        2: {'yes': 'positive', 'no': 'negative'},
        3: {'yes': 'positive', 'no': 'negative'}
    }

    return criterions, tree
```
\end{lstlisting}

When we test the SST-2 performance based on a generated Cola decision tree in Table \ref{table:sst2-tree}, we use the following prompt:

\begin{lstlisting}[
    basicstyle=\scriptsize,
    linewidth=\columnwidth,
    breaklines=true
]
Write a Python function that constructs a decision tree according to the given examples that can generate the correct label of the given classification task.

### Available APIs(shared for all tasks):

# Returns whether the hypothesis in entailed by the premise.
def entailment(hypothesis, premise, model, tokenizer):
    proposition = f'{hypothesis} is entailed by {premise}.'
    inputs = tokenizer(proposition,  return_tensors="pt", truncation=True, padding=True, max_length=128)
    outputs = model(**inputs)['logits'][0]
    ent_label = int(outputs[0] > outputs[2])
    if ent_label == 1:
        return 'yes'
    else:
        return 'no'

# Use the constructed decision tree to predict the label of the sentence.
def tree_predict(sentence, criterions, tree, model, tokenizer):
    node = tree['root']
    while node not in POSSIBLE_CLASSES:
        ent_label = entailment(criterions[node], sentence, model, tokenizer)
        node = tree[node][ent_label]
    return node

### Task: Grammar correctness classification
### Possible classes: ['accpetable', 'unacceptable']

### Define possible classes
POSSIBLE_CLASSES = ['accpetable', 'unacceptable']

### Decision Tree Logic:
- If verbs are not correctly constructed, the sentence is immediately labeled as unacceptable.
- If verbs are correct:
    The tree then checks if the sentence has correct punctuation
    - If incorrect, label the sentence as unacceptable
    - If correct:
        The next criterion to be assessed is the subject-verb agreement.
        - If subject and verb disagree, label the sentence as unacceptable.
        - If they agree, check for sentence fragments.
            - If the sentence is a fragment, label it as unacceptable.
            - If it is not a sentence fragment, label the sentence as acceptable.

### Python code for the decision tree:

```python
def get_decision_tree(sentence, model, tokenizer):
    # Step 1: define criterions of the decision tree
    criterions = {
        'correct_verbs': 'The verbs are correctly constructed in the sentence',
        'correct_punctuation': 'The sentence is punctuated correctly',
        'subject_verb_agreement': 'The subject and verb agree in the sentence',
        'no_sentence_fragments': 'The sentence is not a fragment',
    }

    # Step 2: define the balanced decision tree for this classification task
    tree = {
        'root': 'correct_verbs',
        'correct_verbs': {'yes': 'correct_punctuation', 'no': 'unacceptable'},
        'correct_punctuation': {'yes': 'subject_verb_agreement', 'no': 'unacceptable'},
        'subject_verb_agreement': {'yes': 'no_sentence_fragments', 'no': 'unacceptable'},
        'no_sentence_fragments': {'yes': 'acceptable', 'no': 'unacceptable'}
    }

    return criterions, tree
```
\end{lstlisting}

The input format of target tasks is
\begin{lstlisting}[
    basicstyle=\scriptsize,
    linewidth=\columnwidth,
    breaklines=true
]
### Task: Grammar correctness classification
### Possible classes: [acceptable, unacceptable]
\end{lstlisting}
\clearpage

\section{Implementation Details for Task-Specific Strategies}
\label{appendix-2-fewshot-symbolic}
We detail the few-shot chain-of-thought (CoT) and program-of-thought (PoT) prompting under the task-specific setting in Tables \ref{table:symbolic-reasoning} and \ref{table:strategyqa}:
\begin{itemize}
    \item \textbf{Tracking Shuffled Objects (7).} We use the same 3-shot examples as used by previous work for both task-specific CoT and PoT. The three examples are related to Tracking Shuffled Objects (3) and the models need to learn from demonstrations and generalize to seven objects test cases. The difference between CoT and PoT lies on the format of intermediate reasoning: CoT adopts natural language as the reasoning chains while we transform the thought process into concise Python code for PoT. 
    \item \textbf{Dyck Language.} We cite the results of CoT from LATM \citep{cai2023large} and transform the reasoning steps of the 3-shot examples used by previous chain-of-thought work into Python code for PoT. In order to evaluate the generalization ability of program-of-thought prompting, we try to avoid directly giving generally applicable Python program that can be used for all test instances.
    \item \textbf{Word Sorting.} We cite the results of CoT from LATM \citep{cai2023large} and transform the reasoning steps of the 3-shot examples used by previous chain-of-thought work into Python code for PoT. Since the task can be effectively resolved by just few lines of code, i.e., read in the given input and use \texttt{sorted()} function, e.g.,
    \begin{lstlisting}[
    basicstyle=\scriptsize,
    linewidth=\columnwidth,
    breaklines=true
]
# Python code, return ans
words = ['oven', 'costume', 'counterpart']
ans = " ".join(sorted(words))
\end{lstlisting}
    it can be regarded that the generally applicable tool is already given in the input prompt.
    \item \textbf{Chinese Remainder Theorem.} We cite the results of CoT from LATM \citep{cai2023large}. We build the in-context examples (3-shot) with the first three successful instances of task-general PoT as we construct the Python reasoning chains from the generated programs of task-general PoT with GPT-4. Indeed, for this complicated task, the provided program in the demonstration can also be regarded as a generally applicable tool. That's a main reason why task-specific PoT can obtain 100\% accuracy on this benchmark.
    \item \textbf{Scheduling Meeting.} We cite the results of CoT from LATM \citep{cai2023large}. We build the in-context examples (3-shot) with the first three successful instances of task-general PoT as we construct the Python reasoning chains from the generated programs of task-general PoT with GPT-4. However, unlike giving the ``ground-truth" problem solving tool for Chinese Remainder Theorem, the provided Python reasoning chains can only derive the correct answer for each specific demonstration question but can not be directly applied to all scenarios due to the complexity of the problem. We hope to compare this setup with Chinese Remainder Theorem and evaluate the performance of task-specific PoT on complicated tasks through different in-context learning demonstrations.
    \item \textbf{GSM-Hard.} We use the same 8-shot examples as used by previous work on GSM8K dataset for CoT GSM-Hard. For PoT, we adopt the 9-shot examples on GSM8K dataset from program-of-thought \citep{chen2022program} containing concise Python code as reasoning chains.
    \item \textbf{StrategyQA.} We remove 1 example that appears in the development set from the 6-shot demonstration of previous work \citep{lyu2023faithful} for CoT. As PoT is not designed and applied for natural language question answering task, we did not reproduce task-specific PoT results on StrategyQA benchmark.

\end{itemize}
\clearpage

\section{Evaluation Prompts for VicunaQA}
We have two metrics for VicunaQA. The first metric assesses the level of details and biases to long responses generated by GPT-4, while the other metric does not ask for details.

\subsection{Evaluation prompt asking for details.}
\label{appendix:prompt-detail}

\begin{lstlisting}[
    basicstyle=\scriptsize,
    linewidth=\columnwidth,
    breaklines=true
]
prompt = f'''[Question]\n{ques_str}
        
[The Start of Assistant 1's Answer]\n{gpt4_res}
\n[The End of Assistant 1's Answer]

[The Start of Assistant 2's Answer]\n{target_res}
\n[The End of Assistant 2's Answer]

[System]
We would like to request your feedback on the performance of two AI assistants in response to the user question displayed above.\nPlease rate the helpfulness, relevance, accuracy, level of details of their responses. Each assistant receives an overall score on a scale of 1 to 10, where a higher score indicates better overall performance.\nPlease first output a single line containing only two values indicating the scores for Assistant 1 and 2, respectively. The two scores are separated by a space. In the subsequent line, please provide a comprehensive explanation of your evaluation, avoiding any potential bias and ensuring that the order in which the responses were presented does not affect your judgment.'''
\end{lstlisting}

\subsection{Evaluation prompt not asking for details.}
\label{appendix:prompt-no-detail}

\begin{lstlisting}[
    basicstyle=\scriptsize,
    linewidth=\columnwidth,
    breaklines=true
]
prompt = f'''[Question]\n{ques_str}
        
[The Start of Assistant 1's Answer]\n{gpt4_res}
\n[The End of Assistant 1's Answer]

[The Start of Assistant 2's Answer]\n{target_res}
\n[The End of Assistant 2's Answer]

[System]
We would like to request your feedback on the performance of two AI assistants in response to the user question displayed above.\nPlease rate the relevance and accuracy of their responses. Each assistant receives an overall score on a scale of 1 to 10, where a higher score indicates better overall performance.\nPlease first output a single line containing only two values indicating the scores for Assistant 1 and 2, respectively. The two scores are separated by a space. In the subsequent line, please provide a comprehensive explanation of your evaluation, avoiding any potential bias and ensuring that the order in which the responses were presented does not affect your judgment. Do not bias on either longer or shorter answers.'''
\end{lstlisting}

\subsection{Calculation of Length Bias}
\label{appendix:lb}
Suppose we have $N$ evaluation cases, each receiving 2 candidate responses. A GPT-4 scorer decides the winner between the candidates. $a$ stands for the number of cases where a candidate response with more tokens wins. The length bias is calculated by

\begin{equation}
lb = |\frac{a}{N} - 0.5| * 2
\end{equation}

\end{document}